\documentclass{article}

\usepackage{microtype}
\usepackage{graphicx}
\usepackage{booktabs} 

\usepackage{hyperref}

\usepackage[accepted]{icml2024}

\usepackage{amsmath}
\usepackage{amssymb}
\usepackage{mathtools}
\usepackage{amsthm}

\newcommand{\blue}[1]{{\color{black}#1}}

\usepackage{url}
\usepackage{algorithm}
\usepackage[]{mdframed}
\usepackage{caption}
\usepackage{subcaption}
\usepackage{soul}
\usepackage{wrapfig}
\usepackage{amssymb}
\usepackage{mathrsfs}
\usepackage{bbm}
\usepackage[capitalize,noabbrev]{cleveref}

\theoremstyle{plain}

\theoremstyle{definition}

\theoremstyle{remark}

\usepackage[textsize=tiny]{todonotes}

\icmltitlerunning{Mastering Robot Manipulation with Multimodal Prompts through Pretraining and Multi-task Fine-tuning}

\begin{document}

\twocolumn[
\icmltitle{Mastering Robot Manipulation with Multimodal Prompts through Pretraining and Multi-task Fine-tuning}



\icmlsetsymbol{intern}{*}

\begin{icmlauthorlist}
\icmlauthor{Jiachen Li}{ucsb,intern}
\icmlauthor{Qiaozi Gao}{amazon}
\icmlauthor{Michael Johnston}{amazon}
\icmlauthor{Xiaofeng Gao}{amazon}
\icmlauthor{Xuehai He}{ucsc,intern}
\icmlauthor{Hangjie Shi}{amazon}
\icmlauthor{Suhaila Shakiah}{amazon}
\icmlauthor{Reza Ghanadan}{amazon}
\icmlauthor{William Yang Wang}{ucsb}
\end{icmlauthorlist}

\icmlaffiliation{ucsb}{Department of Computer Science, University of California, Santa Barbara, USA}
\icmlaffiliation{ucsc}{Department of Computer Science, University of California, Santa Cruz, USA}
\icmlaffiliation{amazon}{Amazon AGI}

\icmlcorrespondingauthor{Jiachen Li}{jiachen\_li@ucsb.edu}

\icmlkeywords{Machine Learning, Embodied AI, Multimodal Learning, ICML}

\vskip 0.3in
]



\printAffiliationsAndNotice{$^*$ Work being done during internship at Amazon AGI.}  

\begin{abstract}
Prompt-based learning has been demonstrated as a compelling paradigm contributing to large language models' tremendous success (LLMs). Inspired by their success in language tasks, existing research has leveraged LLMs in embodied instruction following and task planning. 
In this work, we tackle the problem of training a robot to understand multimodal prompts, interleaving vision signals with text descriptions.
This type of task poses a major challenge to robots' capability to understand the interconnection and complementarity between vision and language signals. In this work, we introduce an effective framework that learns a policy to perform robot manipulation with multimodal prompts from multi-task expert trajectories. Our methods consist of a two-stage training pipeline that performs inverse dynamics pretraining and multi-task finetuning. To facilitate multimodal understanding, we design our multimodal prompt encoder by augmenting a pretrained LM with a residual connection to the visual input and model the dependencies among action dimensions. Empirically, we evaluate the efficacy of our method on the VIMA-BENCH~\citep{jiang2023vima} and establish a new state-of-the-art (10\% improvement in success rate). Moreover, we demonstrate that our model exhibits remarkable in-context learning ability. Project page: \url{https://midas-icml.github.io/}.
\end{abstract}

\begin{figure*}[t]
     \centering
     \includegraphics[width=\textwidth]{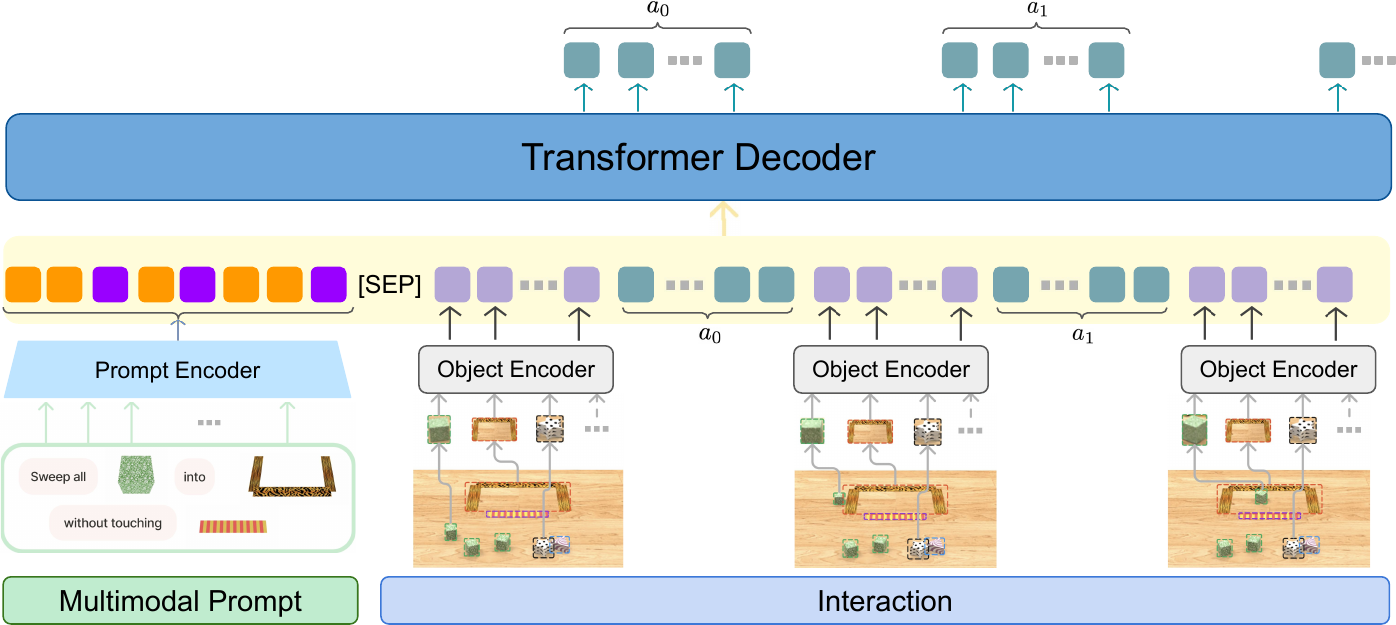}
    \caption{\textbf{Model Architecture of our MIDAS}. Our model adopts a decoder-only architecture. The multimodal prompt embeddings are concatenated with history observation and action tokens. We model each action dimension as an individual token and predict them auto-regressively.}
    \label{fig:model_arc}
\end{figure*}

\section{Introduction}\label{sec:intro}

The unprecedented advancement of large language models (LLM)~\citep{brown2020language,OpenAI2023GPT4TR,team2023gemini,chowdhery2022palm,anil2023palm,chung2022scaling,touvron2023llama} has stimulated rapid development of building instruction-following agents~\citep{lynch2020language,saycan2022arxiv,driess2023palm,guhur2023instruction,huang2022language}. By leveraging LLM's remarkable zero-shot generalizability, various research initiatives~\cite{saycan2022arxiv,huang2022language,huang2022inner} have developed powerful action planners to parse language instructions into a sequence of sub-goals. A prominent example is the SayCan~\citep{saycan2022arxiv}, which employs PALM~\citep{chowdhery2022palm} to transform abstract task descriptions into actionable step-by-step plans. 

However, relying solely on language instructions can be inefficient for describing intricate task details. For instance, directing a household robot to tidy out a living room is more straightforward with a combination of language and visual cues than using language alone. Also, when learning new tasks, words simply cannot convey as much information as video demonstrations~\citep{dasari2021transformers}. In addition, human communication is inherently multimodal, often combining speech with expressive gestures and demonstrations~\citep{drijvers2023multimodal}. Therefore, we are motivated to enhance a robot's comprehension of multimodal task prompts that interleave text and images.


Training a robot to interpret multimodal prompts involves several challenges. The vision signals in the prompt can represent target objects, delineate a specific sub-goal, or offer in-context demonstrations. The robot must understand the underlying transition dynamics suggested by the multimodal prompts before tackling the overall task objective. This requires the robot to infer state transitions from language instructions, and deducing actions from image demonstrations, a concept known as inverse dynamic prediction. Furthermore, it is crucial for the robot to focus on critical visual details, such as the orientation of an object shown in the image, as this can significantly influence its action prediction.

Matching object appearance with textual representation can be achieved by multi-task imitation learning on a diverse set of tasks~\citep{jiang2023vima}. However, imitation learning falls short in teaching robots to predict inverse dynamics, as future observations are often masked out when training to predict actions from current and history observations. To overcome this challenge, we introduce a two-stage training pipeline consisting of inverse dynamic pretraining and multi-task finetuning (FT). Our pretraining strategy first converts any robot trajectory into a motion-following task and then trains the robot to recover the action sequences given the observed image sequence. To capture fine-grained visual information, we design our multimodal prompt encoder by augmenting a pretrained LM with a residual connection (RC) adding from the input visual tokens to the encoded embeddings of the LM.

\autoref{fig:model_arc} provides an overview of our model, which adopts a decoder-only architecture~\citep{radford2018improving}. Specifically, we model each action dimension as an individual action token and predict them auto-regressively to capture dependencies among different dimensions. We dub our method as \textbf{M}ulti-modal \textbf{I}nverse \textbf{D}ynamics \textbf{A}gent\textbf{S} (MIDAS). Empirically, we evaluate our method on the VIMA-BENCH~\citep{jiang2023vima} and establish a new state-of-the-art, outperforming VIMA by ${\sim}10\%$ on all 4 evaluation protocols of VIMA-BENCH. \blue{Our improvement is even more obvious on challenging tasks of VIMA-BENCH, where we achieved performance improvements of 31.8\% on Task 5, 86.3\% on Task 9, 41.0\% on Task 10, and 19.3\% on Task 17 (\autoref{tab:main-result})}. Furthermore, we showcase our multi-task policy's superior in-context learning ability by modifying the original VIMA-BENCH and \blue{designing extra tasks with in-context robot demonstration in the prompt}. \blue{We emphasize this is novel, as simultaneously equipping a robot with multi-task and in-context learning abilities has not been extensively explored in prior research.}

Our contributions can be summarized as follows:
\begin{itemize}
    \item Introduction of the two-stage MIDAS training framework, which establishes a new state-of-the-art on VIMA-BENCH~\citep{jiang2023vima}.
    \item \blue{An effective multimodal prompt encoder that can capture visual and textual details.}
    \item \blue{Equipping a multi-task robot with the in-context learning ability. To the best of our knowledge, this has not been extensively explored in prior research.}
\end{itemize}

\section{Preliminary}
\textbf{Problem Definition}
We consider the problem of learning a multimodal prompt-conditioned policy $\pi: \mathcal{P}\times\Omega\rightarrow\mathcal{A}$ that maps the multimodal prompt $q\in\mathcal{P}$ and the history trajectory $\omega_t = \left(o_0, a_0, o_1, \ldots, a_{t-1}, o_t \right)\in\Omega$ to the two-pose action primitive~\citep{zeng2021transporter} $a_t = (\mathcal{T}_{\text{initial}}, \mathcal{T}_{\text{target}})\in\mathcal{A}\subseteq\mathcal{R}^{N_a}$, where $o_t\in\mathcal{O}$ denotes the visual observation at timestep $t$ and $N_a$ denotes the number of action dimensions.
\begin{equation}
    \begin{aligned}
        \pi(q, \omega_t) = \pi\left(q, o_0, a_0, o_1, \ldots, a_{t-1}, o_t \right) \\
        \rightarrow a_t = (\mathcal{T}_{\text{initial}}, \mathcal{T}_{\text{target}})\in\mathcal{A}\subseteq\mathcal{R}^{N_a}
    \end{aligned}
\end{equation}
The action space $\mathcal{A}$ consists of primitive motor skills like ``pick and place" and ``push". For the ``pick and place" primitive, $\mathcal{T}_{\text{initial}}$ and $\mathcal{T}_{\text{target}}$ defines the space of pick and place pose, respectively. For ``push", they define the space of the starting and ending pose of push. The multimodal prompt describes the task goal by interleaving texts and images.

In this paper, we aim to learn a multi-task policy $\pi_\theta$ parameterized by $\theta$ from a dataset $\mathcal{D} = \{\zeta_1, \ldots, \zeta_N\}$ with $N$ expert demonstration. Each training sample $\zeta_i = (q^i, \omega^i)$ contains the expert trajectory $\omega^i = \left(o^i_0, a^i_0, o^i_1, \ldots, a^i_{T-1}, o^i_T \right)$ corresponding to the multimodal task prompt $q_i$.

\begin{figure}[t]
     \centering
     \begin{subfigure}[b]{0.495\linewidth}
         \centering
         \includegraphics[width=\linewidth]{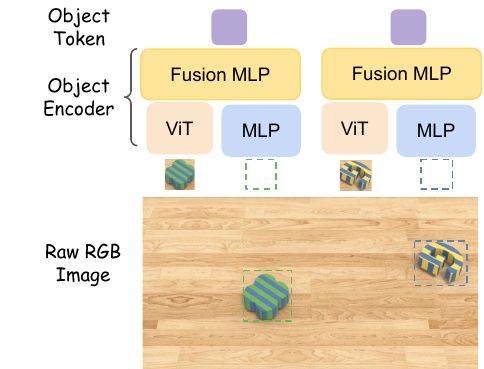}
         \caption{Object Encoder}
         \label{fig:obj_encoder}
     \end{subfigure}
     \hfill
     \begin{subfigure}[b]{0.495\linewidth}
         \centering
         \includegraphics[width=\linewidth]{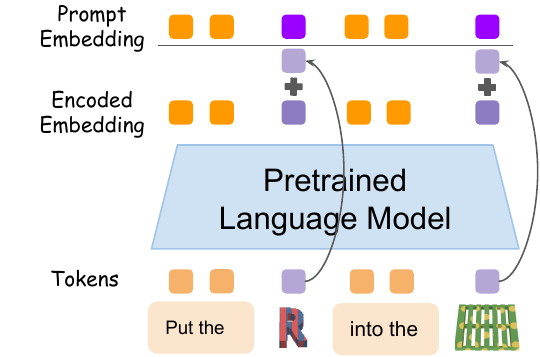}
         \caption{Multimodal Prompt Encoder}
         \label{fig:prompt_encoder}
     \end{subfigure}
  \caption{(a) \textbf{Object Encoder} proposed in VIMA consists of a ViT~\citep{dosovitskiy2020image} that extracts visual embedding from cropped object images and a MLP that encodes bounding boxes. The two embeddings are concatenated before passing through a Fusion MLP to get the object tokens. (b) \textbf{Multimodal Prompt Encoder} adds a RC from the input object tokens to the pretrained LM output.}\label{fig:encoders}
\end{figure}

\textbf{VIMA policy} \cite{jiang2023vima} propose the VisuoMotor Attention (VIMA) agent to solve robot manipulation from multimodal prompts with a Transformer~\citep{vaswani2017attention} Encoder-Decoder architecture. It encodes the task prompts that interleave images and texts with a pretrained LM by following the practice of Frozen~\citep{tsimpoukelli2021multimodal}. Its autoregressive action decoding is conditioned on the prompt embedding via cross attention layers that alternate with the causal self-attention. Instead of directly operating on the raw RGB images, VIMA adopts the object-centric representation by cropping objects from both prompt and observation images and forming them as a sequence of object tokens with pixel coordinate information as shown in \autoref{fig:obj_encoder}. Notably, VIMA predicts each action dimension independently and trains its model via behavior cloning with the loss function for a trajectory with $T$ steps given by
\begin{equation}
    \begin{aligned}
        L(\theta) & = -\sum^{T-1}_{t=0}\log\pi_\theta(a_t|q, \omega_t) \\
        & = -\sum^{T-1}_{t=0}\sum^{N_a-1}_{n=0}\log\pi_\theta(a^n_t|q, \omega_t).
    \end{aligned}
\end{equation}
We build our policy upon the VIMA policy. However, we model the dependencies among different action dimensions~\citep{giuliari2021transformer,alphastarblog} and decode each dimension autoregressively. We detail our motivation in Sec. \ref{sec:indep_action_decode} and demonstrate its empirical benefit in Sec. \ref{sec:exp}.

\begin{figure*}[t]
  \makebox[\textwidth]
    {
        \begin{subfigure}{0.26\paperwidth}
            \center
            \includegraphics[width=\linewidth]{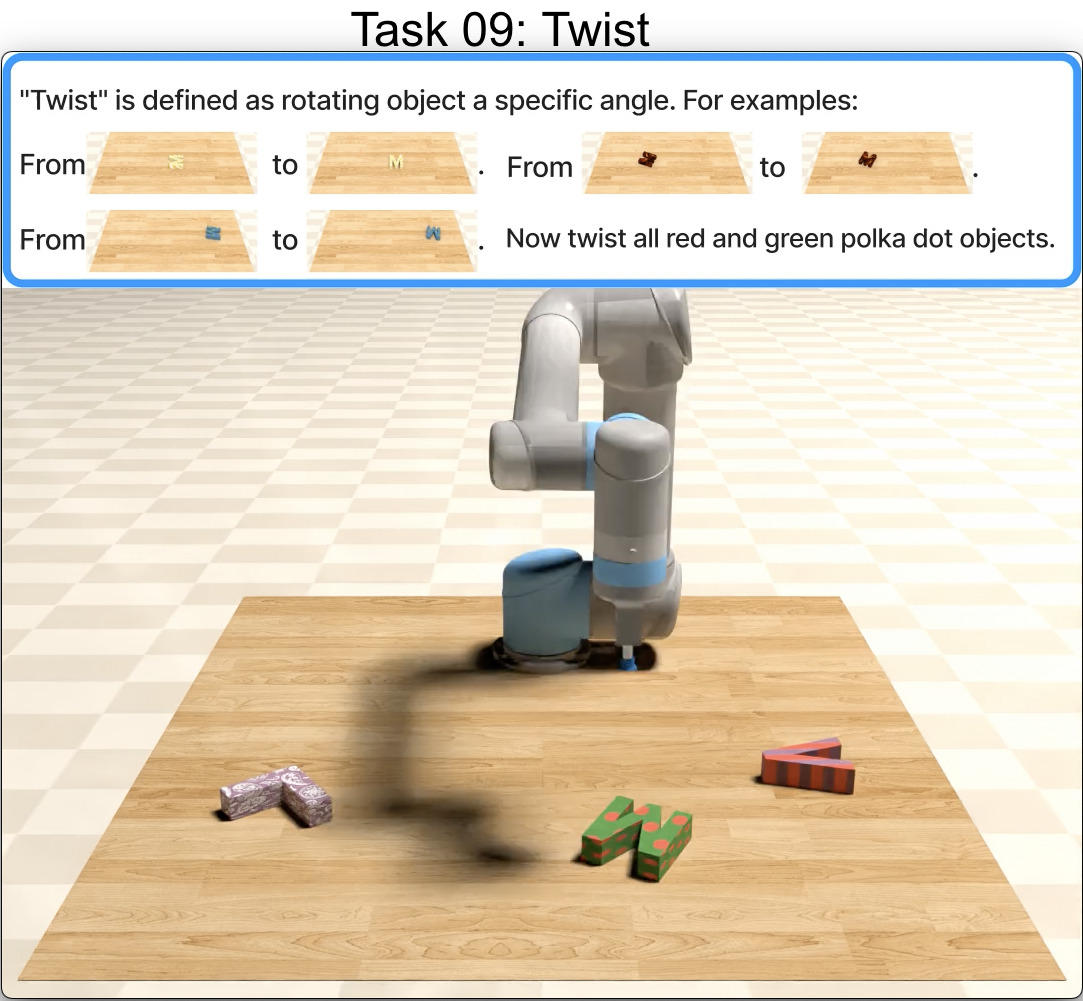}
        \end{subfigure}
        \begin{subfigure}{0.26\paperwidth}
            \center
            \includegraphics[width=\linewidth]{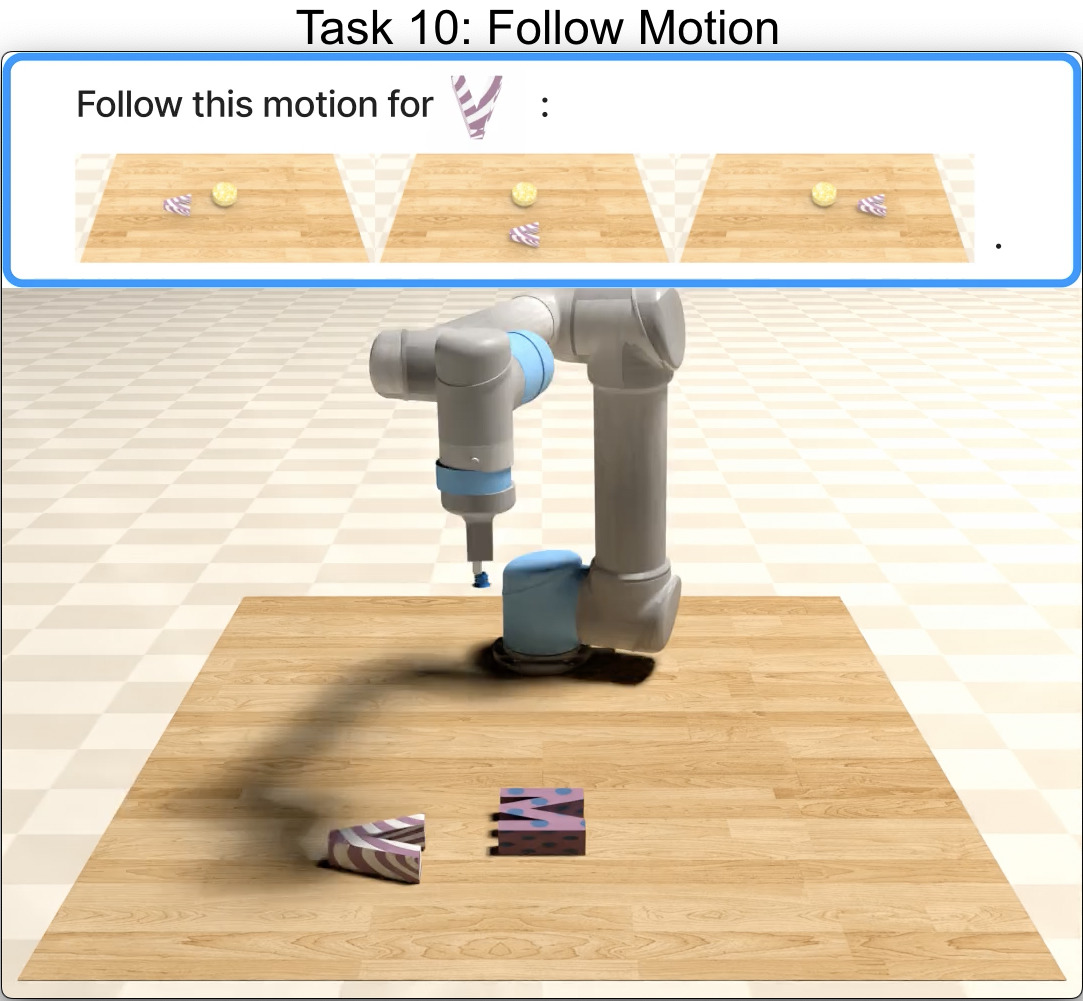}
        \end{subfigure}

        \begin{subfigure}{0.26\paperwidth}
            \center
            \includegraphics[width=\linewidth]{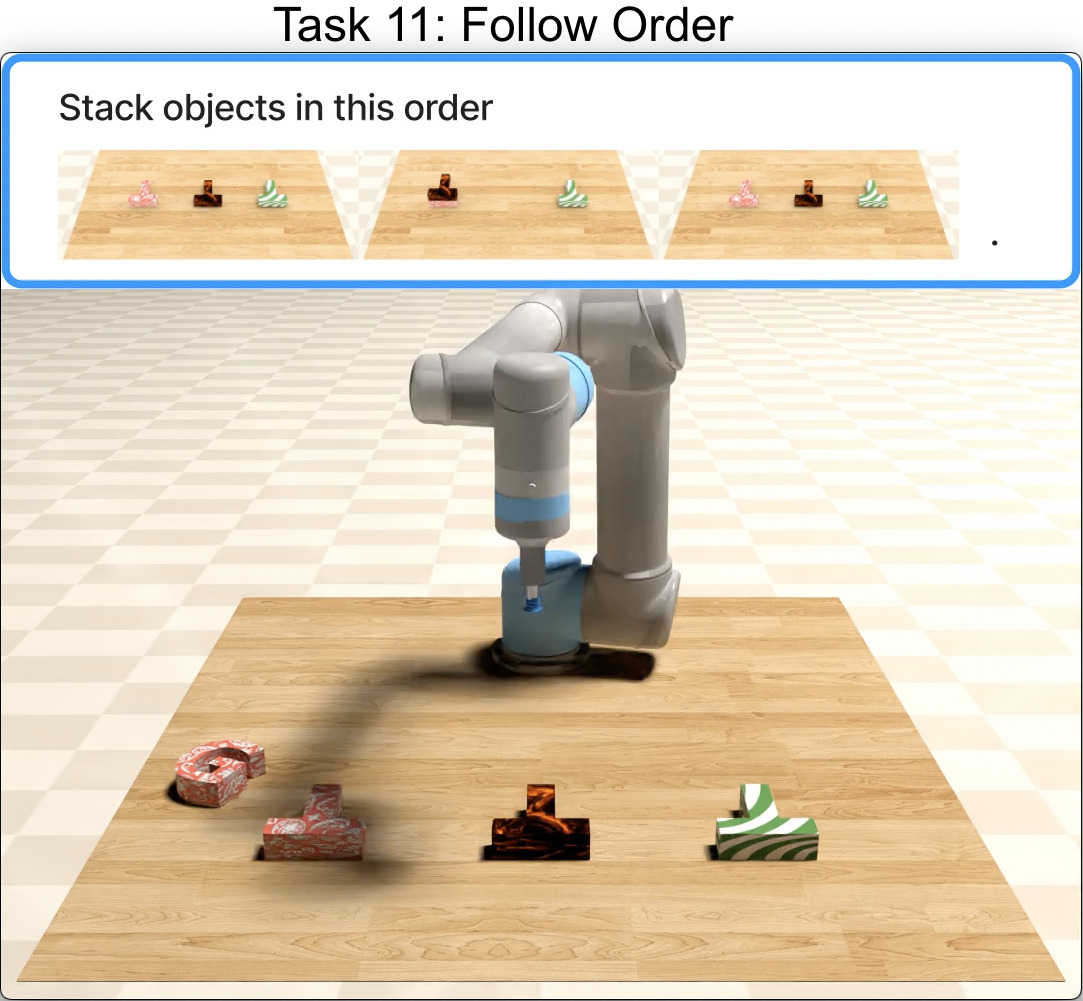}
        \end{subfigure}
    }
    \makebox[\textwidth]
    {
        \begin{subfigure}{0.26\paperwidth}
            \center
            \includegraphics[width=\linewidth]{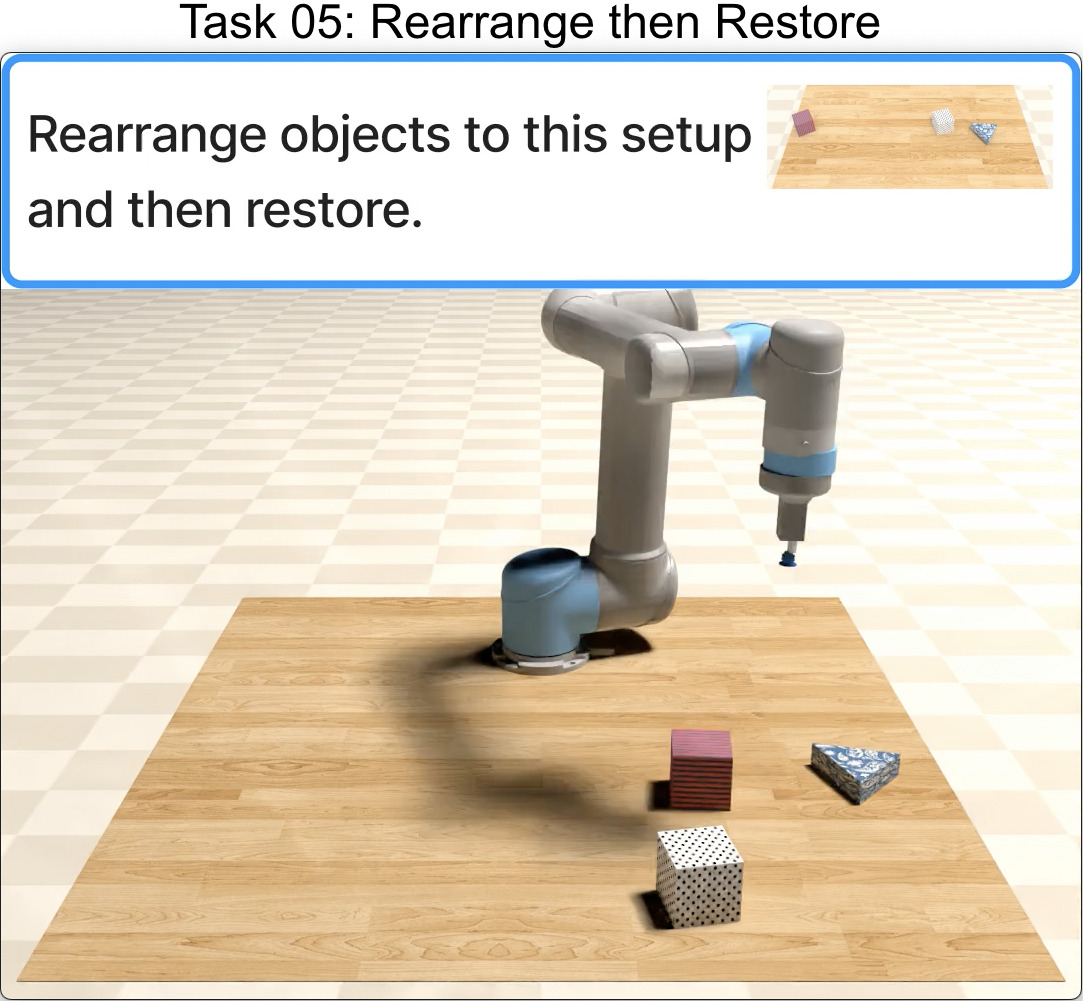}
        \end{subfigure}
        
        \begin{subfigure}{0.26\paperwidth}
            \center
            \includegraphics[width=\linewidth]{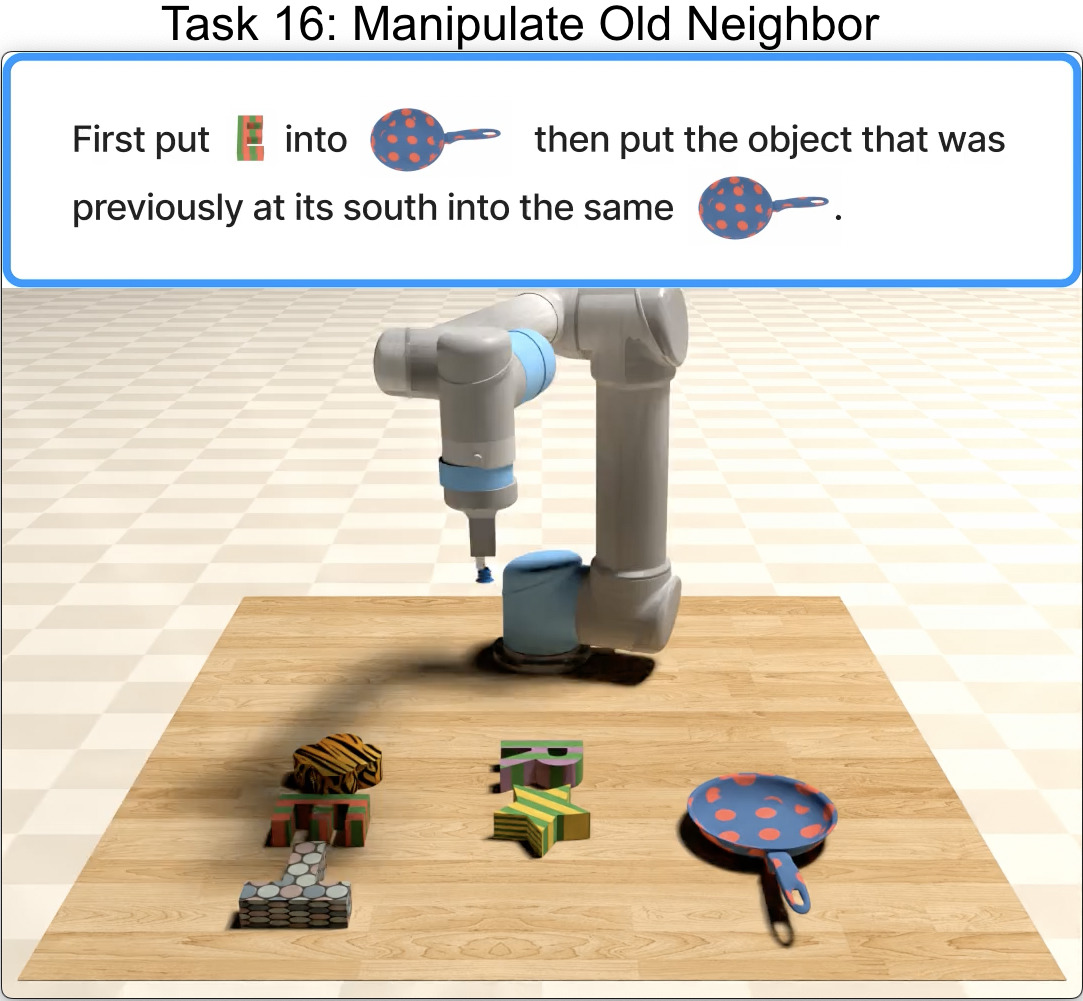}
        \end{subfigure}

        \begin{subfigure}{0.26\paperwidth}
            \center
            \includegraphics[width=\linewidth]{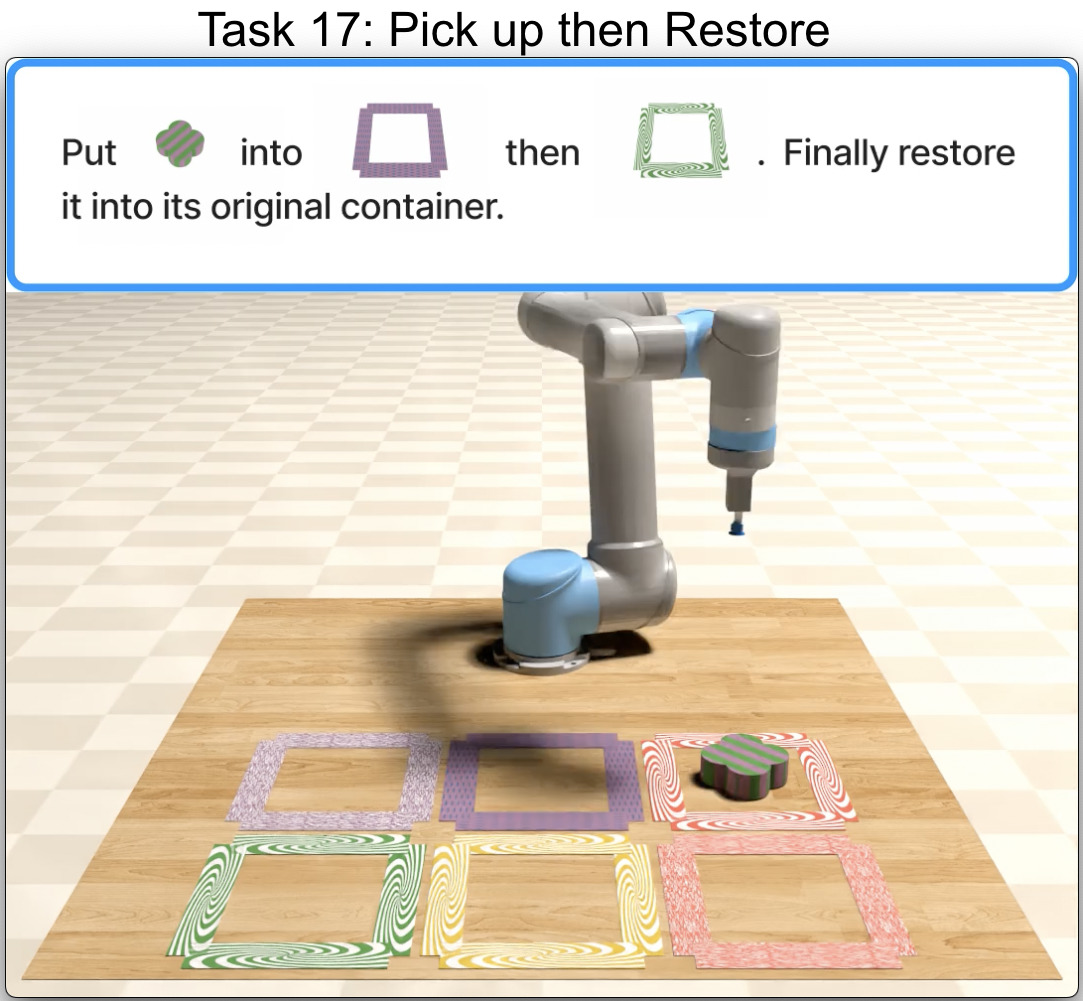}
        \end{subfigure}
    }
  \caption{Task samples from the VIMA-BENCH. We refer readers to Appendix B of the VIMA paper~\citep{jiang2023vima} for detailed task description.
  }
\label{fig:vima-tasks}
\end{figure*}

\textbf{VIMA-BENCH}~\citep{jiang2023vima} is built on top of the Ravens~\citep{zeng2021transporter,shridhar2023perceiver} simulator and contains 17 types of tabletop manipulation tasks. \autoref{fig:vima-tasks} shows 6 representative tasks from the VIMA-BENCH. Each task type can instantiate thousands of individual task instances by combining various textures and objects. Specifically, each task instance defines a multimodal prompt that interleaves texts and images and the type of end-effector $\in\{\text{suction cup, spatula}\}$. The suction cup corresponds to the primitive motor skill ``pick and place" while spatula corresponds to ``wipe". At each time step, the agent receives RGB images rendered from both frontal and top-down views and predicts the initial and target pose of its end effector.

VIMA-BENCH establishes a four-level protocol to evaluate progressively stronger generalization, ranging from placement generalization (L1), combinatorial generalization (L2), novel object generalization (L3) and novel task generalization (L4). Expert demonstration are provided for 13 tasks as the training data, with 50K trajectories per task. The other 4 tasks are included into the L4 task suite.

\section{Methods}

We introduce our MIDAS framework that learns a multi-task policy to perform robot manipulation with multimodal prompt. We propose a two-stage training pipeline that includes inverse dynamic pretraining (Sec. \ref{sec:train_pipeline}) followed by multi-task FT. To capture fine-grained visual information, we design our multimodal prompt encoder by augmenting a pretrained LM with a residual connection to the input object token (Sec. \ref{sec:prompt_encode}). Moreover, we model each action dimension as an individual action token and autoregressively decodes each dimension (Sec. \ref{sec:indep_action_decode}). Sec. \ref{sec:alg_sum} summarizes our training framwork, with an overview of our model architecture given in \autoref{fig:model_arc}.

\subsection{Pretraining Task: Inverse Dynamics Prediction}\label{sec:train_pipeline}
\begin{figure}[ht]
    \centering
     \includegraphics[width=\linewidth]{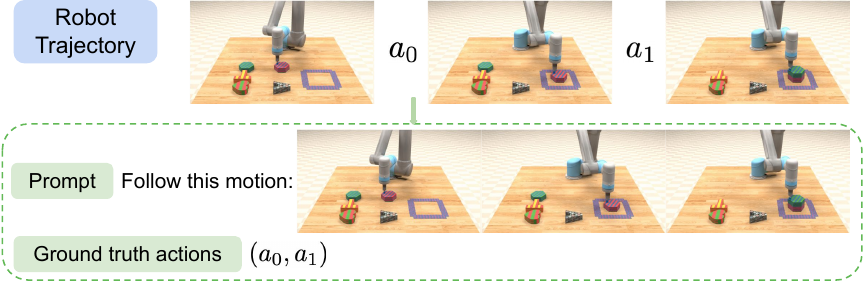}
    \caption{Given the any sequence of robot trajectory, we can always formulate a motion following task that requires the agent to replicate the demonstration trajectory.}
    \label{fig:inverse_dp}
\end{figure}


As mentioned in Sec. \ref{sec:intro}, images in the prompt can depict object appearance, appearances, outline the sub-goals and success criteria of a task, or serve as in-context task demonstrations. To decipher this underlying task information and learn from in-context examples, a robot needs to understand the transition dynamics illustrated in a sequence of images. For instance, the robot should be able to infer the action sequence required to transition from its current state to the target goal state.

In other words, the agent needs proficiency in inverse dynamics prediction. Given a sequence of observations $(o_0, \ldots, o_{T})$, the robot should learn to infer the corresponding action sequence $(a_0, \ldots, a_{T-1})$. However, the skill cannot be directly acquired by imitating multi-task trajectories, as future observations are often masked out when predicting actions with current observations. 

To tackle the dilemma, we make a novel observation that every robot trajectory itself can be reformulated into a motion following task. As shown in \autoref{fig:inverse_dp}, given any sequence of robot trajectory $\omega_T = \left(o_0, a_0, o_1, \ldots, a_{T-1}, o_T \right)$, we can always create a task with the prompt $q_{pretrain} = (\textit{Follow this motion: }o_0, \ldots, o_{T})$ and ground-truth actions $(a_0, \ldots, a_{T-1})$, leading to the following pretraining loss
\begin{equation}\label{eq:pretrain_loss}
     L_{\text{pretrain}}(\theta) = -\sum^{T-1}_{t=0}\log\pi_\theta(a_t|q_{pretrain}; \omega_t)
\end{equation}
We can paraphrase $q_{pretrain}$ with a LM~\citep{OpenAI2023GPT4TR,team2023gemini} to enhance its language diversity during pretraining. For simplicity, we leave it for future research.


\subsection{Multi-modal Prompt Encoding}\label{sec:prompt_encode}
To capture visual and textual information from the multimodal prompt, VIMA proposes to encode both the visual and language tokens in the prompt with a pretrained LM (T5-base) following the practice of Frozen~\citep{tsimpoukelli2021multimodal}. While LLM has demonstrated a tremendous success across various fields with superior generalizability~\citep{li2022pre}, our early experiments reveal that this encoding strategy often fails to capture some fine-grained visual information, e.g., the rotation angle of an object (Task 09, \autoref{fig:vima-tasks}). We hypothesize it is because the pretrained LM has never been trained on visual data.

To overcome this challenge, we propose to augment the pretrained LM by adding a residual connection (RC) 
from the input visual tokens to the encoded embeddings , as shown in \autoref{fig:prompt_encoder}. The intuition is that by directly adding the original visual tokens to the embeddings produced by the pretrained LM, we can retain more detailed visual information that might be lost during the encoding process. Our experiments in Sec. \ref{sec:exp} validate this intuition, showing that the inclusion of the RC significantly improves performance across different tasks.




\subsection{Modeling the Dependency Among Each Action Dimension}\label{sec:indep_action_decode}
\begin{figure}[t]
     \centering
     \includegraphics[width=\linewidth]{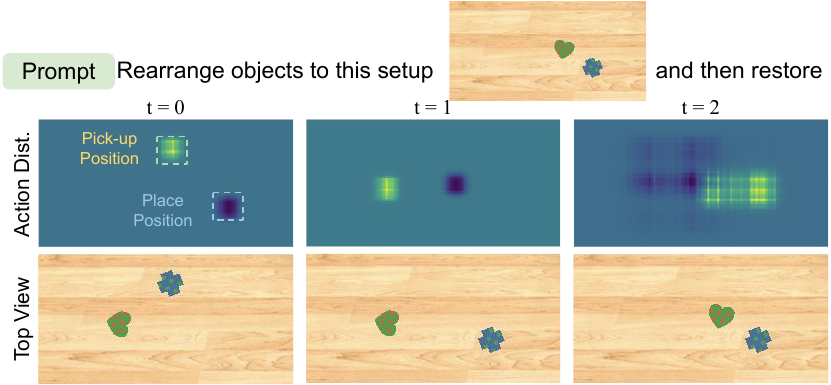}
    \caption{At $t = 2$, the robot should move either the heart or the cross block. As the policy predicts each action dimension independently, different dimensions do not consistently manipulate the same object, resulting in a task failure.}    
    \label{fig:confused_dist}
\end{figure}

Recall that the robot action is defined by the initial pose $\mathcal{T}_{\text{initial}}$ and target pose $\mathcal{T}_{\text{target}}$ of the end effector. Intuitively, $\mathcal{T}_{\text{target}}$ should depend on $\mathcal{T}_{\text{initial}}$. And thus independently predicting each action dimension can be problematic. Consider the example in \autoref{fig:confused_dist}, the robot is tasked to first rearrange the objects to a desired arrangement and then restore them to original setup. When the robot begins to restore at $t=2$, it has the option to move either the heart or the cross block. As the policy predicts each action dimension independently, the dimensions associated with the pick-up pose do not align consistently with one specific object. Consequently, the distribution of pick-up position assigns significant probability to both object locations. Similarly, the placement position distribution allocates probability to both objects' target positions. When sampling actions from this distribution, the robot may either miss picking up an object or misplace it, leading to a task failure.


Therefore, we opt to model the dependency among action dimensions by modeling each dimension as a single token and decode each token autoregressively as shown in \autoref{fig:model_arc}. And thus, the multi-task imitation loss function can be reformulated into
\begin{equation}\label{eq:imitate_loss}
    \begin{aligned}
         L_{\text{Imitation}}(\theta) & = -\sum^{T-1}_{t=0}\bigg(\log\pi_\theta(a^0_t|q, \omega_t) \\
         & + \sum^{N_a-1}_{n=1}\log\pi_\theta(a^{n}_t|q, \omega_t, a_t^{0}, \ldots, a_t^{n-1})\bigg).
    \end{aligned}
\end{equation}
That is, the distribution for each action dimension should be conditioned on the other action dimensions that have already been decoded.

\subsection{Algorithm Summary}\label{sec:alg_sum}
To this end, we have introduced our pretraining strategies and model design. To learn our multi-task policy $\pi_{\theta}$, we assume the access to a dataset  $\mathcal{D} = \{\zeta_1, \ldots, \zeta_N\}$ with $N$ expert demonstration. First, we pretrain $\pi_{\theta}$ by minimizing $L_{\text{pretrain}}(\theta)$ over $N_{\text{pretrain}}$ iterations. Subsequently, we perform multi-task fine-tuning to minimize $L_{\text{Imitation}}(\theta)$. The pseudo-codes (Algorithm \ref{alg}) and detailed hyper-parameters (HP) are available in Appendix \ref{sec:pseudo-codes}.


\section{Experimental Results on VIMA-BENCH}\label{sec:exp}

\blue{This section aims to evaluate whether our model design and training pipeline enhance the zero-shot generalization of the learned model. We conduct experiments on the VIMA-BENCH (Sec. \ref{sec:exp-vima}) and carry out extensive ablation studies (Sec. \ref{sec:ablation}).
}

\subsection{Standard Evaluation on the VIMA-BENCH}\label{sec:exp-vima}
\begin{table*}
    \caption {We compared our methods with baseline approaches on the VIMA-BENCH across all four evaluation levels. ``Avg" represents the average success rate for all tasks within an evaluation level. To determine the success rate for each method, we sampled 200 episodes from every task. Due to limited space, we report the success rate for four representative tasks in this table. Full results can be found in Appendix \ref{sec:appen-full-results}. Our methods significantly outperform baseline methods and establish a new state-of-the-art performance on the VIMA-BENCH.}\label{tab:main-result}
    \centering
    \resizebox{\linewidth}{!}{
        \begin{tabular}{l|lllllllllllllllllllllllllll}
            \toprule
              & \multicolumn{6}{c}{L1} & \multicolumn{6}{c}{L2} & \multicolumn{6}{c}{L3} & \multicolumn{2}{c}{L4}    \\ 
            \cmidrule[0.5pt]{2-6} \cmidrule[0.5pt]{8-12} \cmidrule[0.5pt]{14-18}  \cmidrule[0.5pt]{20-21}
            Method  & Avg & T5 & T9 & T16 & T17 &
                    & Avg & T5 & T9 & T16 & T17 &
                    & Avg & T5 & T9 & T16 & T17 &
                    & Avg & T10 \\
            \midrule
             Gato         & 57.0 & 44.5 & 14.0 & 43.0 & 1.5 & 
                          & 53.9 & 46.0 & 10.5 & 42.0 & 1.0 & 
                          & 45.6 & 36.0 & 17.0 & 41.5 & 0.0 &
                          & 13.5 & 0.0 \\
             Flamingo     & 47.2 & 41.0 & 3.0  & 38.0 & 2.0 & 
                          & 47.1 & 43.0 & 4.5 & 40.0 & 1.0 & 
                          & 42.1 & 36.5 & 6.0 & 45.5 & 0.5 &
                          & 11.1 & 0.0 \\
             GPT          & 47.9 & 45.0 & 8.0  & 33.0 & 1.0 & 
                          & 47.4 & 43.0 & 10.5 & 34.0 & 3.0 & 
                          & 42.6 & 32.0 & 5.0 & 37.5 & 0.0 &
                          & 12.1 & 0.5 \\
            \midrule
             VIMA         & 87.2 & 65.0 & 13.5 & 88.0 & 77.0 & 
                          & 87.0 & 61.0 & 12.5 & 87.5 & 77.5 & 
                          & 84.0 & 63.0 & 12.0 & $\mathbf{58.5}$ & 78.0 &
                          & 49.6 & 0.0 \\
             Gato OBJ     & 87.5 & 62.0 & 17.0 & 92.5 & 80.5 & 
                          & 87.5 & 62.5 & 16.0 & 91.5 & 80.0 & 
                          & 84.4 & 65.5 & 15.5 & 46.5 & 87.5 &
                          & 49.6 & 0.0 \\
            \midrule
            \textbf{Ours}\\
             w/o Pretrain & 91.6 & 88.0 & 20.5 & 93.0 & $\mathbf{98.0}$ & 
                          & 91.8 & 87.0 & 23.5 & 92.0 & $\mathbf{98.0}$ & 
                          & 88.3 & 90.0 & 20.5 & 50.5 & $\mathbf{99.5}$ &
                          & 49.1 & 0.0 \\

             w/ Pretrain  & $\mathbf{97.8}$ & $\mathbf{94.0}$ & $\mathbf{100}$ & $\mathbf{94.0}$ & 96.5 & 
                          & $\mathbf{97.9}$ & $\mathbf{96.5}$ & $\mathbf{100}$ & $\mathbf{93.0}$ & {96.0} & 
                          & $\mathbf{93.4}$ & $\mathbf{94.0}$ & $\mathbf{97.0}$ & {47.0} & $\mathbf{98.0}$ &
                          & $\mathbf{59.1}$ & $\mathbf{41.0}$ \\
    
             \bottomrule
        \end{tabular}
    }
\end{table*}

We compare our methods with various baselines from the VIMA paper~\citep{jiang2023vima} on the VIMA-BENCH. All baseline methods only conduct multi-task imitation learning without pretraining. We directly report results for Gato~\citep{reed2022generalist}, Flamingo~\citep{alayrac2022flamingo} and GPT~\citep{radford2018improving} from the VIMA paper. Notably, these three methods directly operate on the raw image observation. In contrast, VIMA, Gato OBJ and our methods adopt an object-centric representation. The Gato OBJ policy is constructed by replacing VIMA's encoder-decoder architecture with a decoder-only architecture~\citep{radford2018improving}. And the difference between our policy and Gato OBJ is that we augments the pretrained LM with a RC and model each action dimension as an individual action token. As we do not focus on the visual understanding part of general robot control, we assume the access to the ground truth instance segmentation masks provided by the VIMA-BENCH for all methods with an object-centric representation. And the results of VIMA and Gato OBJ are reproduced by us.

\autoref{tab:main-result} presents the results by following VIMA-BENCH's 4-level evaluation protocols. Due to the limited space, we only report the individual task success rates for representative tasks on which different methods exhibit a significant performance difference. Avg denotes the task success rate across all tasks from an evaluation level. Appendix \ref{sec:appen-full-results} includes full evaluation results with individual task success rate. We can observe that our methods already outperforms all baseline methods even without pretraining, particularly on Task 5 (\emph{Rearrange the Restore}) and Task 17 (\emph{Pick up then Restore}), demonstrating the effectiveness of our multimodal prompt encoder and the importance of modeling the dependencies between initial and target pose of the action. With pretraining, the performance of our methods improves significantly, especially on the difficult Task 9 (\emph{Twist}) and Task 10 (\emph{Follow Motion}). As shown in \autoref{fig:vima-tasks}, \emph{Twist} requires the robot to first deduct the target rotation angles from the in-context examples before operating on the correct objects described by text. Similarly, \emph{Follow Motion} requires the robot to deduce the actions corresponding to the image sequence in the prompt and apply them to the same object in robot's current observation.
Without pretraining, models have to learn the skills for inverse dynamics prediction solely from the multi-task data, lacking enough supervision.


\subsection{Ablation Studies}\label{sec:ablation}

\begin{figure*}[t]
     \centering
     \includegraphics[width=\textwidth]{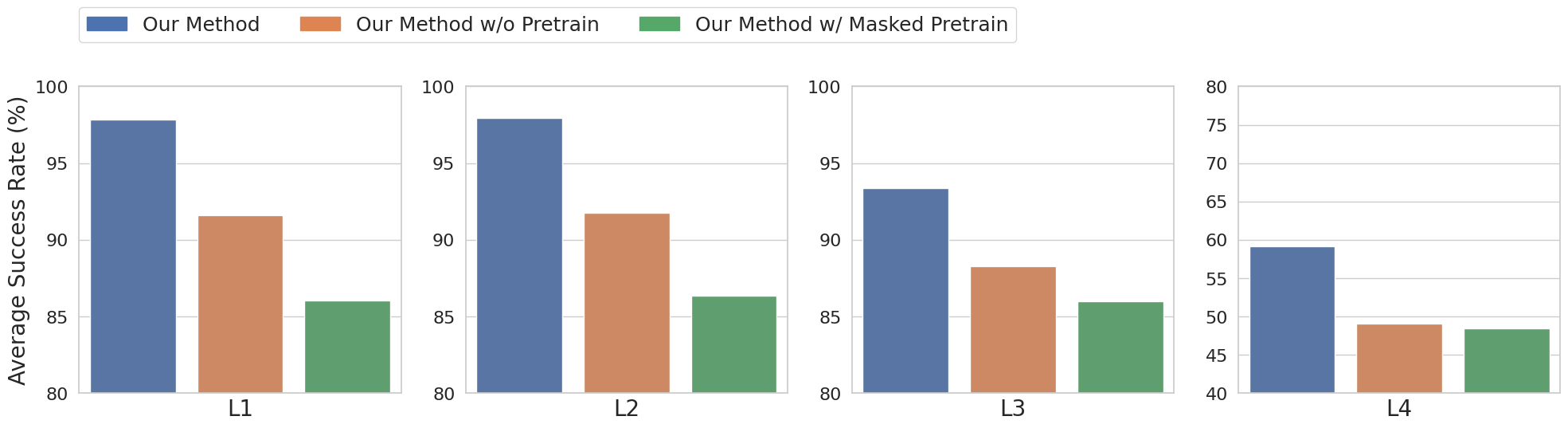}
    \caption{Ablation study on the pretraining strategy. We show that the BERT-style pretraining strategy (Our Method w/ Masked Pretrain) that performs masked action modeling does not benefit the learning of a multi-task policy to understand multimodal prompts.}
    \label{fig:ablate_pretrain}
\end{figure*}

\begin{figure*}[t]
     \centering
     \includegraphics[width=\textwidth]{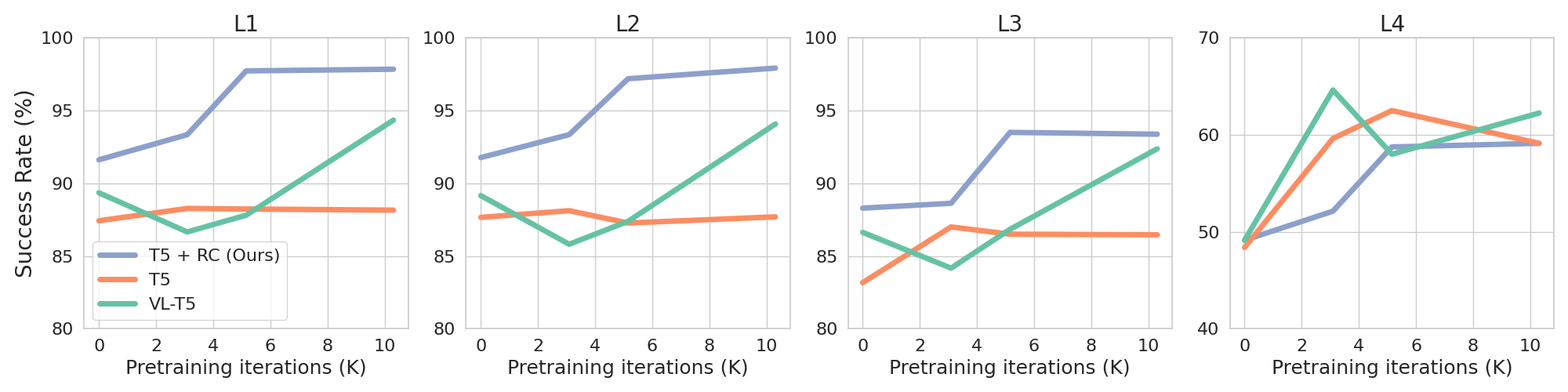}
    \caption{
    Ablation on the prompt encoder. We compare the performance of our methods with different prompt encoders. Our proposed T5 + RC prompt encoder that augments a pretrained T5 with a residual connection (RC) to the input visual tokens achieves a higher computational efficiency by requiring less pretraining iterations to reach a decent performance on L1, L2, and L3.
    }
    \label{fig:ablate_prompt_encoder}
\end{figure*}

\begin{figure*}[t]
     \centering
     \includegraphics[width=\textwidth]{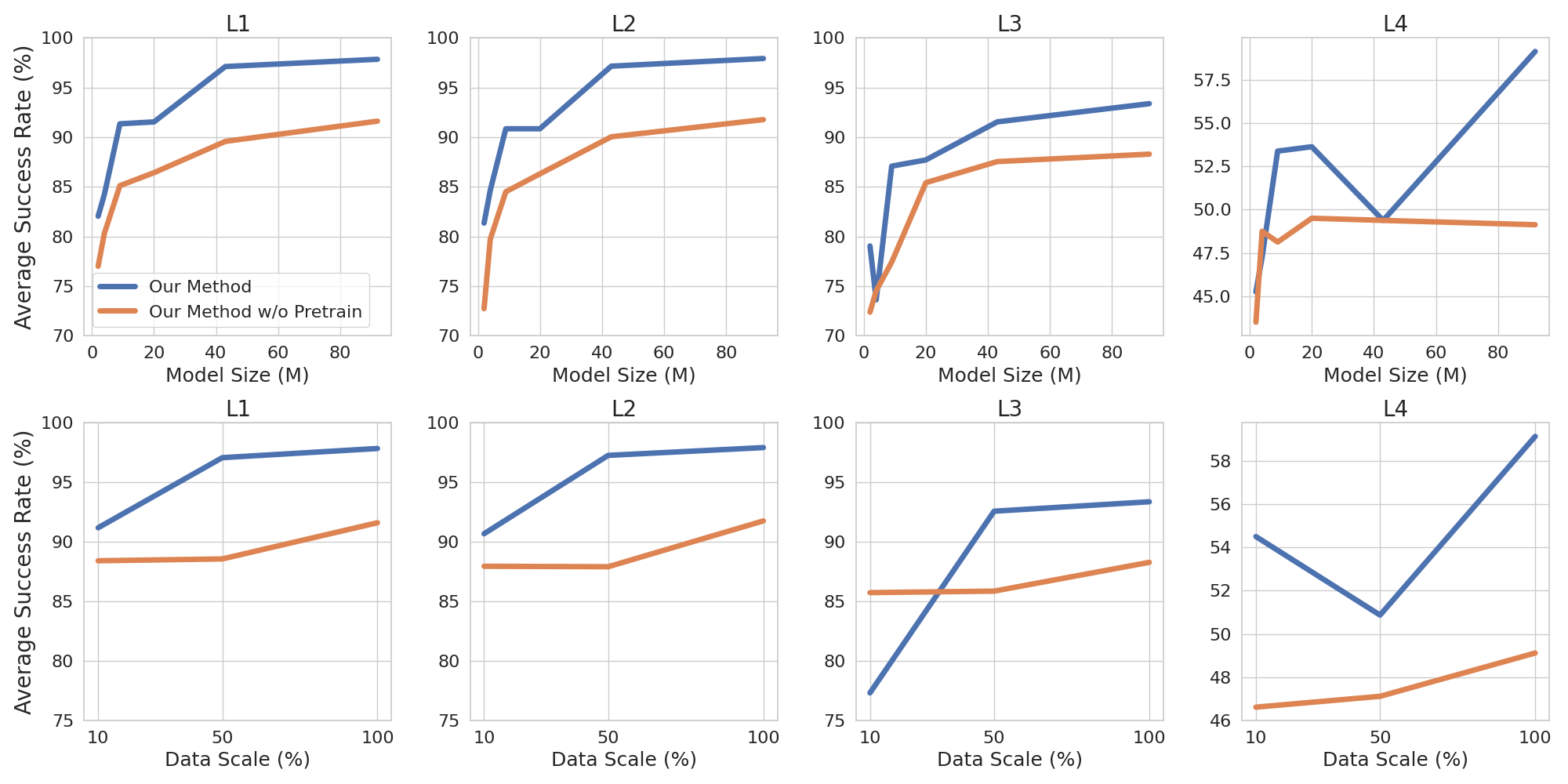}
    \caption{Ablation on model and data sizes. \textit{Top}: For model sizes ranging from 2M to 92M, our pretraining can always learn a representation that leads to better multitask performance. \textit{Bottom}: Model size is fixed to 92M. The benefit of pretraining increases as we increase training data size.}
    \label{fig:ablate_data_model_size}
\end{figure*}

We conduct extensive experiments to study how our model design and training pipeline impacts the robot manipulation, focusing on the effectiveness of our pretraining strategy and prompt encoding. We also examine the impact of data scaling and model size. Appendix \ref{sec:appen-full-results} presents individual task success rate for all methods and further ablate the decoder-only architecture of our model. Appendix \ref{sec:add-exp} studies the effectiveness of the number of gradient steps.

\textbf{Pretraining Strategy}.
\autoref{fig:ablate_pretrain} compared our pretraining strategy with a BERT-style masking prediction method~\citep{devlin2018bert}, which still performs the task of inverse dynamics prediction. Specially, we modify the decoding mask of the transformer to allow its attention to all future observation but mask all prompt and future action tokens. However, this pretraining strategy does not benefit the downstream multitask learning, as it does not explicitly train the model to reason the image sequences presented in the prompt.

\textbf{Multimodal Prompt Encoding}.
Recall that our multimodal prompt encoder (T5 + RC) augments a pretrained LM (T5-Base~\citep{raffel2020exploring}) with a RC to the input visual tokens. To investigate its efficacy, we compare its performance with two variants that respectively adopt a pretrained T5 and VL-T5~\citep{cho2021unifying} to encode the multimodal prompt. Note that T5 is pretrained on pure text data while VL-T5 is pretrained on both vision and language data. As shown in \autoref{fig:ablate_prompt_encoder}, our method achieves overall better performance and computational efficiency by requiring less pretraining iterations. This remains true even with additional gradient steps (Appendix \ref{sec:more-iter-vl-t5}). The comparison between the performance of T5 and VL-T5 shows that a pretrained encoder that better understands input visual tokens can benefit more from our pretraining phase.

\textbf{Model \& Data Scalability}.
\autoref{fig:ablate_data_model_size} illustrates the performance of our methods in relation to variations in model and data size. As the model size is scaled from 2M to 92M, we maintain a constant prompt encoder and exclude it from the parameter count, adhering to VIMA practices. Conversely, while adjusting the data size, the model remains fixed at 92M, and performance is evaluated using 10\%, 50\%, and 100\% of the data available in VIMA-BENCH. Notably, the enhancements derived from our pretraining remain evident in low parameter regime. Additionally, a larger dataset correspondingly amplifies the performance gains achieved through our pretraining techniques.

\section{Evaluating the In-context Learning Ability}\label{sec:in-context-exp}

\begin{table*}[t]
    \caption{Evaluating the in-context learning capability. We hold out \emph{Twist} and \emph{Follow Order} from the training data.}
    \centering
    \scalebox{1.0}{
        \begin{tabular}{l|ccc|c}
            \toprule
            Task &  T9: Twist  & T10: Follow Motion & T11: Follow Order & Overall \\
            \midrule
            Our Method & $\mathbf{26.5\%}$ & $\mathbf{74.0\%}$ & 8.0 \%     & $\mathbf{36.2 \%}$ \\
            \midrule
            Our Method w/o Modified FT & 10.0\%  & 43.5 \% & $\mathbf{16.5\%}$  & 23.3\% \\
            \midrule
            Our Method w/ Pretrain Only & 8.0 \% & 2.0\% & 15.5 \%  & 8.5 \% \\
            \midrule
            Our Method w/o Pretrain & 1.5 \%  & 0.5 \% &  0.0\%  & 0.7 \% \\
            \bottomrule
        \end{tabular}
    }
    \label{tab:icl-results}
\end{table*}

Previous experiments in Sec. \ref{sec:exp} already demonstrate the superior generalizability of our methods on L1, L2 and L3, which differs from the training tasks in object placement, combination and types. When exposed to novel tasks from L4, we expect our framework imbues the agent with a human-like intuition to learn from in-context examples. This expectation holds even if none of the training tasks explicitly present few-shot demonstrations within their prompts.

\blue{To access whether our model can effectively utilize in-context examples to tackle novel tasks}, we modify the original VIMA-BENCH by carefully constructing a new set of L4 tasks, ensuring each of the L4 tasks contain in-context examples in the prompt. Specifically, we hold out \emph{Twist} and \emph{Follow Order} from the training tasks, combining them with \emph{Follow Motion} to form the new L4 task suites. The first row of \autoref{fig:vima-tasks} showcases samples from these designated tasks.

As L4 tasks contain novel scenes/objects that does not exist in the training data, we leverage data augmentation during pretraining phase to improve model generalizability. Additionally, we propose \emph{Modified FT} that randomly replace the object image in the prompt with text description provided by the VIMA-BENCH during multi-task finetuning. At inference time, we edit the prompt of \emph{Twist} and \emph{Follow Order} to make them closer to the pretraining prompt without adding extra task information. Appendix~\ref{sec:appen-in-context-exp} provides detailed experiment setup.

As shown in \autoref{tab:icl-results}, our method considerably outperforms baseline methods for the \emph{Twist} and \emph{Follow Motion} without decreasing its performance on L1, L2 and L3 (shown in Appendix~\ref{sec:appen-in-context-exp}). \emph{Twist} requires the model to first infer the rotation angle from prompt image sequence and then identify the target object described with text. While the imitation-learned policy (Our Method w/o Pretrain) shows limited performance on these tasks, our pretrained policy (Our Method w/ Pretrain Only) exhibits some capability, particularly in \emph{Follow Order} which does not necessitate understanding object descriptions. However, it has difficulties with \emph{Twist} and \emph{Follow Motion} because it has never trained to tackle the visual and textual object. In contrast, the multi-task FT phase helps the model to understand diverse multimodal prompts and solicit its ability to translate action sequences derived from in-context examples to target objects. This is akin to the improvement seen in pretrained language models' instruction following abilities due to instruction-fining, as highlighted by~\citep{sanh2021multitask,ouyang2022training,wei2021finetuned}. Moreover, our Modified FT significantly improves model's grounding capability, contributing to a remarkable performance increase in \emph{Twist} and \emph{Follow Motion}.

Appendix \ref{sec:add_icl_exp} provides additional results, where we design 4 new tasks with in-context examples in the prompt to solidify our findings.

\section{Related Work}

\textbf{Multi-Task Pretraining via Sequence Modeling.} The development of the Transformer architecture~\citep{vaswani2017attention} paved the way for large-scale pretraining, which has become a standard practice to enable better generalization across different domains~\citep{brown2020language,chen2021pix2seq,radford2021learning,devlin2018bert,lu2022unified,li2023offline,li2024reward}. Specifically, these models employ the \textit{sequential modeling}~\citep{sutskever2014sequence} techniques to capture temporal dependencies in the data. By training on massive web-scale data, the trained models demonstrate emergent behaviors~\citep{brown2020language,chowdhery2022palm,touvron2023llama}, e.g., the ability to perform in-context learning. While multi-task pretraining has been extensively employed in natural language processing (NLP) and computer vision (CV), its applications in robotic systems are also gaining increasing attention~\citep{driess2023palm,brohan2022rt,brohan2023rt,radosavovic2023robot}. In our work, we pretrain our model by converting diverse robot trajectories into inverse dynamics prediction tasks, facilitating our in-context learning and multi-task performance. 

\textbf{Multimodal Learning.} The field of multimodal learning, which focuses on integrating data from various modalities, has seen remarkable advancements~\citep{radford2021learning, wang2022image, jaegle2021perceiver}. Flamingo, for instance, trains a model to generate textual completion based on multimodal prompts~\citep{alayrac2022flamingo}. The Perceiver framework~\citep{jaegle2021perceiver} offers an adaptable method to process structured input and output. Moreover, Gato~\citep{reed2022generalist} introduces a versatile agent proficient in NLP, CV, and robotics. Our research tackles robot manipulation given interleaved image and text task prompt. Similarly, MUTEX~\citep{shah2023mutex} learns a policy to tackle task prompts from multiple modalities (image, video, text, and speech). However, each task in MUTEX is defined in a single modality. Thus, their task prompts do not interleave different modalities.

\textbf{Inverse Dynamics Modeling (IDM) for Representation Learning.} IDM has proved to be an effective approach for learning from high-dimensional demonstration data. Training the model on an IDM task of predicting the agent's actions given the high-dimensional observations allows effective learning of a feature space that represents only the information relevant to the actions~\citep{brandfonbrener2023inverse}. \cite{pathak2017curiosity} uses IDM to generate intrinsic reward signals with self-supervision for efficient exploration. \cite{efroni2021provable, lamb2022guaranteed} use a multi-step inverse dynamics model to enable representation learning robust to exogenous information. Most recently, \cite{baker2022video, venuto2023multi, thomas2023plex} use IDM for data-efficient multi-task pre-training on complex sequential decision-making domains. Our method leverages IDM to facilitate robot's in-context learning capability and its understanding on the  transition dynamics.

\section{Conclusion}
In this paper, we introduce our MIDAS framework that trains a robot to tackle multimodal prompts. The pretraining phase trains the agent to perform inverse dynamics prediction, facilitating robot's understanding of transition dynamics. To capture fine-grained visual information from the prompt images, we augment a pretrained LM with a RC to the object token. We further model the dependency among different action dimensions. Empirically, we establish a new state-of-the-art on the VIMA-BENCH and also demonstrate the in-context learning capability of our learned policy.

\section{Limitations}\label{sec:limitation}
\textbf{Limited Task Complexity}. To the best of our knowledge, VIMA-BENCH is the only existing benchmark that considers multimodal task prompts that interleave text and images. While our MIDAS framework already establishes a SOTA performance on VIMA-BENCH, we further expand the VIMA-BENCH by designing four new tasks to strengthen our in-context learning results in Appendix \ref{sec:add_icl_exp}.

\textbf{Limited Motion Primitives}. Our experiments mainly focus on table-top robot manipulation with the pick and place and push motion primitives on the VIMA-BENCH. However, our MIDAS framework is designed to be general-purpose and can support any motion primitive $(\mathcal{T}_\text{initial}, \mathcal{T}_\text{target}) \in \mathcal{A}$ that can be parameterized by the initial pose $\mathcal{T}_\text{initial}$ and target pose $\mathcal{T}_\text{target}$ of the end effector. For example, it is possible to extend our MIDAS framework to support low-level action spaces like joint-torque control with minimal modifications.

\section*{Impact Statement}
Our work could be a transformative step in the realm of human-robot collaboration. Our study introduces a framework that enhances robots' understanding of multimodal prompts that interleaves visual and textual inputs in a seamless and effective manner. The approach of inverse dynamics pretraining and multi-task finetuning, as demonstrated in our work, not only sets a new benchmark in robotic manipulation tasks but also opens up vast opportunities for more intuitive and efficient human-robot interactions in various settings. The significant improvement in robotic task success rates and in-context learning abilities signifies a potential paradigm shift in how robots can be integrated into workplaces, offering new dimensions of assistance, precision, and adaptability. This advancement promises to revolutionize industries, enhance productivity, and pave the way for more dynamic human-robot partnerships, fundamentally reshaping the future landscape of work.
\clearpage
\bibliography{example_paper}
\bibliographystyle{icml2024}

\clearpage

\appendix
\onecolumn
\begin{center}
	{\LARGE {Appendix}}
\end{center}

\section{Individual Task Success Rate of Different Methods}\label{sec:appen-full-results}
In this section, we support all experimental results in Sec. \ref{sec:exp} of our main paper with individual task success rates for all four levels of evaluation protocol. Specifically, the results for \autoref{tab:main-result} and the ablation on Pretraining strategies can be found in Table \ref{tab:full-result-L1}, \ref{tab:full-result-L2}, \ref{tab:full-result-L3} and \ref{tab:full-result-L4}. We conduct a fine-grained analysis of the performance gain achieved by Our Method w/ Pretrain and Our Method w/o Pretrain against the VIMA policy across the 4 evaluation protocols.
\begin{enumerate}
    \item For tasks with motion demonstrations (T9, T10, T11), Our Method w/ Pretrain (90.6\%) improves over the VIMA policy (47.1\%) by 43.5\% while Our Method w/o Pretrain (52.1\%) improves 5.0
    \item For the other tasks without motion demonstrations (14 tasks in total), Our Method w/ Pretrain (93.4\%) improves over the VIMA policy (89.7\%) by 3.7\% while Our Method w/o Pretrain (93.6\%) improves 3.9\%.
\end{enumerate}
The results imply the following fact:
\begin{enumerate}
    \item The performance improvement (43.5\%) made by Our Method w/ Pretrain on tasks with motion demonstrations is significantly higher than the improvement (5.0\%) made by Our Method w/o Pretrain.
    \item The performance improvement (3.7\%) made by Our Method w/ Pretrain on tasks without motion demonstrations is similar to the improvement (3.9\%) made by Our Method w/o Pretrain.

\end{enumerate}

Therefore, we can conclude that the improvement made by our pretraining method DOES NOT overfit on the tasks with motion demonstrations. Our significant improvement on tasks with motion demonstrations is NOT achieved by sacrificing the performance improvements on tasks without motion demonstrations.

\autoref{fig:ablate_prompt_encoder} ablates multimodal prompt encoding is based on the results from Table \ref{tab:prompt-encoder-result-L1}, \ref{tab:prompt-encoder-result-L2}, \ref{tab:prompt-encoder-result-L3} and \ref{tab:prompt-encoder-result-L4}. The results in \autoref{fig:ablate_data_model_size} that ablate model and data sizes are based on the results from Table \ref{tab:model-size-result-L1}, \ref{tab:model-size-result-L2}, \ref{tab:model-size-result-L3}, \ref{tab:model-size-result-L4}, \ref{tab:data-scale-result-L1}, \ref{tab:data-scale-result-L2}, \ref{tab:data-scale-result-L3}, and \ref{tab:data-scale-result-L4}.

Additionally, we further conduct an ablation study on the transformer architecture of our policy by replacing the decoder-only architecture with encoder-decoder architecture (Our Method w/ Encoder-Decoder). Experimental results in Table \ref{tab:full-result-L1}, \ref{tab:full-result-L2}, \ref{tab:full-result-L3} and \ref{tab:full-result-L4} show that this variant does not perform as well as our method on the L1, L2, and L3 tasks, mainly due to its inability to tackle Task 09 (\emph{Twist}) that requires deducting rotation angles from the prompt image sequence (\autoref{fig:vima-tasks}). However, it achieves a superior performance on the L4 Task 10 (\emph{Follow Motion}). We hypothesize that it is due to the limit of model capacity. This policy learns a control policy that predicts its action dependent on the object bounding box, while lacking the capability to capture fine-grained visual information that contains the information of object rotation.

\begin{table}[ht]
    \caption {L1 level generalization results. All methods share the same amount of parameters 92M. Integers in the first row refer to indices of tasks defined in the VIMA paper~\citep{jiang2023vima}}\label{tab:full-result-L1}
    \centering
    \resizebox{\linewidth}{!}{
        \begin{tabular}{l|ccccccccccccc|c}
            \toprule
            Method  & T1 & T2 & T3 & T4 & T5 & T6 & T7 & T9 & T11 & T12 & T15 & T16 & T17 & Overall\\
            \midrule
            VIMA & 100 & 100 & 100 & 100 & 65.0 & 99.5 & 100 & 13.5 & 96.0 & 94.5 & 100 & 88.0 & 77.0 & 87.2\\
            Gato OBJ & 100 & 100 & 100 & 100 & 75.5 & 100 & 100 & 16.5 & 88.5 & 93.0 & 100 & 92.5 & 80.0 & 88.2\\
            \midrule
            Our Method \\
            w/o Pretrain       & 100 & 100 & 100 & 98.5 & 88.0 & 100 & 100 & 20.5 & 100 & 94.0 & 99.0 & 93.0 & 98.0 & 91.6\\
            w/ Pretrain        & 98.5 & 100 & 100 & 99.5 & 94.0 & 100 & 100 & 100 & 100 & 94.0 & 95.5 & 94.0 & 96.5 & 97.8\\
            w/ Masked Pretrain & 100 & 99.5 & 100 & 99.5 & 97.5 & 99.5 & 100 & 17.5 & 74.5 & 94.5 & 97.0 & 42.5 & 96.5 & 86.0\\
            w/ Encoder-Decoder & 100 & 100 & 99.5 & 99.5 & 96.5 & 100 & 100 & 19.5 & 99.5 & 93.5 & 99.0 & 93.0 & 82.5 & 91.0\\
             \bottomrule
        \end{tabular}
    }
\end{table}
\begin{table}[ht]
    \caption {L2 level generalization results. All methods share the same amount of parameters 92M. Integers in the first row refer to indices of tasks defined in the VIMA paper~\citep{jiang2023vima}.}\label{tab:full-result-L2}
    \centering
    \resizebox{\linewidth}{!}{
        \begin{tabular}{l|ccccccccccccc|c}
            \toprule
            Method  & T1 & T2 & T3 & T4 & T5 & T6 & T7 & T9 & T11 & T12 & T15 & T16 & T17 & Overall\\
            \midrule
            VIMA & 100 & 100 & 100 & 99.5 & 61.0 & 100 & 100 & 12.5 & 97.5 & 95.0 & 100 & 87.5 & 77.5 & 87.0\\
            Gato OBJ & 100 & 100 & 100 & 100 & 73.0 & 100 & 100 & 17.5 & 88.5 & 95.0 & 99.0 & 94.0 & 80.5 & 88.3\\
            \midrule
            Our Method \\
            w/o Pretrain & 100 & 100 & 100 & 99.0 & 87.0 & 100 & 100 & 23.5 & 100 & 94.0 & 99.5 & 92.0 & 98.0 & 91.8\\
            w/ Pretrain & 98.5 & 100 & 100 & 99.0 & 96.5 & 99.5 & 100 & 100 & 100 & 95.5 & 95.0 & 93.0 & 96.0 & 97.9\\
            w/ Masked Pretrain & 99.5 & 100 & 100 & 99.5 & 96.5 & 100 & 99.5 & 19.5 & 75.5 & 95.5 & 97.0 & 43.5 & 96.5 & 86.3\\
            w/ Encoder-Decoder & 99.5 & 100 & 99.0 & 99.5 & 96.5 & 100 & 100 & 15.5 & 99.5 & 94.0 & 98.5 & 92.0 & 82.5 & 90.5\\
             \bottomrule
        \end{tabular}
    }
\end{table}
\begin{table}[ht]
    \caption {L3 level generalization results. All methods share the same amount of parameters 92M. Integers in the first row refer to indices of tasks defined in the VIMA paper~\citep{jiang2023vima}.}\label{tab:full-result-L3}
    \centering
    {
        \begin{tabular}{l|cccccccccccc|c}
            \toprule
            Method  & T1 & T2 & T3 & T4 & T5 & T6 & T7 & T9 & T11 & T15 & T16 & T17 & Overall\\
            \midrule
            VIMA & 99.5 & 100 & 100 & 99.5 & 63.0 & 99.5 & 100 & 12.0 & 98.5 & 99.5 & 58.5 & 78.0 & 84.0\\
            Gato OBJ & 99.5 & 100 & 100 & 100 & 72.5 & 97.5 & 100 & 7.5 & 95.0 & 99.5 & 44.5 & 72.0 & 82.3\\
            \midrule
            Our Method \\
            w/o Pretrain & 99.5 & 100 & 100 & 100 & 90.0 & 100 & 100 & 20.5 & 100 & 99.5 & 50.5 & 99.5 & 88.3\\
            w/ Pretrain & 98.0 & 99.0 & 100 & 99.5 & 94.0 & 97.5 & 99.0 & 97.0 & 96.5 & 95.0 & 47.0 & 98.0 & 93.4\\
            w/ Masked Pretrain & 99.0 & 100 & 100 & 100 & 96.5 & 99.5 & 99.0 & 20.5 & 76.5 & 99.0 & 42.0 & 100 & 86.0\\
            w/ Encoder-Decoder & 99.0 & 99.5 & 100 & 98.5 & 95.5 & 99.5 & 98.0 & 20.0 & 100 & 95.0 & 56.0 & 86.0 & 87.2\\
             \bottomrule
        \end{tabular}
    }
\end{table}
\begin{table}[ht]
    \caption {L4 level generalization results. All methods share the same amount of parameters 92M. Integers in the first row refer to indices of tasks defined in the VIMA paper~\citep{jiang2023vima}.}\label{tab:full-result-L4}
    \centering
    {
        \begin{tabular}{l|cccc|c}
            \toprule
            Method & T8 & T10 & T13 & T14 & Overall\\
            \midrule
            VIMA & 98.5 & 0.0 & 0.0 & 100 & 49.6\\
            Gato OBJ & 99.5 & 0.0 & 0.0 & 98.0 & 49.4\\
            \midrule
            Our Method \\
            w/o Pretrain & 97.0 & 0.0 & 0.0 & 99.5 & 49.1\\
            w/ Pretrain & 97.5 & 41.0 & 1.0 & 97.0 & 59.1\\
            w/ Masked Pretrain & 95.0 & 0.0 & 0.0 & 99.0 & 48.5\\
            w/ Encoder-Decoder & 96.5 & 85.5 & 0.0 & 97.0 & 69.8\\
            \bottomrule
        \end{tabular}
    }
\end{table}
\begin{table}[ht]
    \caption {Comparison of the performance of our method with different multimodal prompt encoder on L1 level generalization. All methods share the same amount of parameters 92M. Integers in the first row refer to indices of tasks defined in the VIMA paper~\citep{jiang2023vima}}\label{tab:prompt-encoder-result-L1}
    \centering
    \resizebox{\linewidth}{!}{
        \begin{tabular}{l|l|ccccccccccccc|c}
            \toprule
            Pretrain iter. & Method  & T1 & T2 & T3 & T4 & T5 & T6 & T7 & T9 & T11 & T12 & T15 & T16 & T17 & Overall\\
            \midrule
            0.0K & T5 + RC (Ours) & 100 & 100 & 100 & 98.5 & 88.0 & 100 & 100 & 20.5 & 100 & 94.0 & 99.0 & 93.0 & 98.0 & 91.6\\
            0.0K & T5 & 100 & 99.5 & 98.5 & 99.0 & 84.5 & 100 & 100 & 16.0 & 100 & 94.5 & 98.0 & 49.5 & 97.0 & 87.4\\
            0.0K & VL-T5 & 100 & 100 & 98.5 & 99.5 & 66.0 & 100 & 100 & 18.0 & 100 & 93.0 & 96.5 & 94.0 & 96.0 & 89.3\\
            \midrule
            31K & T5 + RC (Ours) & 100 & 100 & 100 & 98.5 & 85.5 & 100 & 99.0 & 52.0 & 100 & 93.5 & 96.0 & 93.0 & 96.0 & 93.3\\
            31K & T5 & 100 & 100 & 100 & 100 & 98.5 & 100 & 100 & 18.5 & 100 & 93.5 & 97.0 & 43.5 & 96.5 & 88.3\\
            31K & VL-T5 & 100 & 99.5 & 100 & 100 & 70.0 & 99.5 & 100 & 23.5 & 100 & 94.5 & 99.5 & 43.5 & 96.5 & 86.7\\
            \midrule
            52K & T5 + RC (Ours) & 100 & 100 & 100 & 99.0 & 92.0 & 99.5 & 100 & 99.5 & 100 & 95.5 & 95.0 & 93.0 & 97.0 & 97.7\\
            52K & T5 & 100 & 100 & 100 & 98.5 & 98.0 & 100 & 100 & 19.5 & 100 & 94.0 & 98.5 & 42.0 & 96.5 & 88.2\\
            52K & VL-T5 & 100 & 99.5 & 100 & 99.0 & 94.0 & 99.5 & 99.5 & 21.0 & 100 & 94.0 & 95.5 & 43.0 & 96.5 & 87.8\\
            \midrule
            103K & T5 + RC (Ours) & 98.5 & 100 & 100 & 99.5 & 94.0 & 100 & 100 & 100 & 100 & 94.0 & 95.5 & 94.0 & 96.5 & 97.8\\
            103K & T5 & 99.5 & 99.5 & 100 & 97.0 & 98.0 & 99.0 & 99.5 & 22.0 & 100 & 94.0 & 99.5 & 41.0 & 97.0 & 88.2\\
            103K & VL-T5 & 99.5 & 99.0 & 100 & 100 & 97.5 & 99.0 & 100 & 100 & 100 & 93.5 & 98.0 & 43.0 & 97.0 & 94.3\\
             \bottomrule
        \end{tabular}
    }
\end{table}

\begin{table}[ht]
    \caption {Comparison of the performance of our method with different multimodal prompt encoder on L2 level generalization. All methods share the same amount of parameters 92M. Integers in the first row refer to indices of tasks defined in the VIMA paper~\citep{jiang2023vima}}\label{tab:prompt-encoder-result-L2}
    \centering
    \resizebox{\linewidth}{!}{
        \begin{tabular}{l|l|ccccccccccccc|c}
            \toprule
            Pretrain iter. & Method  & T1 & T2 & T3 & T4 & T5 & T6 & T7 & T9 & T11 & T12 & T15 & T16 & T17 & Overall\\
            \midrule
            0.0K & T5 + RC (Ours) & 100 & 100 & 100 & 99.0 & 87.0 & 100 & 100 & 23.5 & 100 & 94.0 & 99.5 & 92.0 & 98.0 & 91.8\\
            0.0K & T5 & 99.5 & 100 & 99.0 & 99.5 & 87.0 & 100 & 99.5 & 20.0 & 100 & 93.5 & 98.5 & 48.0 & 95.0 & 87.7\\
            0.0K & VL-T5 & 100 & 100 & 98.5 & 99.0 & 66.5 & 99.0 & 99.5 & 19.0 & 100 & 94.0 & 96.5 & 92.0 & 95.0 & 89.2\\
            \midrule
            31K & T5 + RC (Ours) & 99.5 & 100 & 100 & 98.5 & 89.5 & 100 & 99.5 & 52.0 & 100 & 94.0 & 92.5 & 92.0 & 96.0 & 93.3\\
            31K & T5 & 99.0 & 100 & 100 & 100 & 97.0 & 99.0 & 99.5 & 22.5 & 100 & 94.5 & 96.0 & 41.5 & 96.5 & 88.1\\
            31K & VL-T5 & 98.0 & 99.5 & 99.5 & 98.5 & 67.5 & 99.0 & 99.5 & 24.0 & 100 & 94.0 & 96.5 & 44.0 & 95.5 & 85.8\\
            \midrule
            52K & T5 + RC (Ours) & 100 & 100 & 100 & 97.5 & 91.0 & 98.5 & 99.5 & 99.5 & 100 & 95.5 & 93.0 & 92.5 & 96.5 & 97.2\\
            52K & T5 & 98.5 & 100 & 100 & 98.0 & 96.5 & 99.0 & 98.5 & 21.5 & 100 & 94.0 & 93.5 & 40.0 & 95.0 & 87.3\\
            52K & VL-T5 & 99.5 & 100 & 100 & 98.5 & 94.0 & 98.5 & 97.5 & 20.0 & 100 & 94.0 & 94.0 & 43.5 & 96.5 & 87.4\\
            \midrule
            103K & T5 + RC (Ours) & 98.5 & 100 & 100 & 99.0 & 96.5 & 99.5 & 100 & 100 & 100 & 95.5 & 95.0 & 93.0 & 96.0 & 97.9\\
            103K & T5 & 100 & 100 & 100 & 97.0 & 97.0 & 100 & 97.0 & 19.5 & 100 & 94.5 & 97.0 & 43.0 & 95.0 & 87.7\\
            103K & VL-T5 & 99.0 & 99.5 & 100 & 99.5 & 97.0 & 99.0 & 99.0 & 99.5 & 100 & 94.5 & 97.5 & 43.0 & 95.5 & 94.1\\
             \bottomrule
        \end{tabular}
    }
\end{table}

\begin{table}[ht]
    \caption {Comparison of the performance of our method with different multimodal prompt encoder on L3 level generalization. All methods share the same amount of parameters 92M. Integers in the first row refer to indices of tasks defined in the VIMA paper~\citep{jiang2023vima}}\label{tab:prompt-encoder-result-L3}
    \centering
    \resizebox{\linewidth}{!}{
        \begin{tabular}{l|l|cccccccccccc|c}
            \toprule
            Pretrain iter. & Method  & T1 & T2 & T3 & T4 & T5 & T6 & T7 & T9 & T11 & T15 & T16 & T17 & Overall\\
            \midrule
            0.0K & T5 + RC (Ours) & 99.5 & 100 & 100 & 100 & 90.0 & 100 & 100 & 20.5 & 100 & 99.5 & 50.5 & 99.5 & 88.3\\
            0.0K & T5 & 98.5 & 98.5 & 100 & 100 & 85.5 & 99.5 & 98.5 & 19.5 & 100 & 98.5 & 42.0 & 57.5 & 83.2\\
            0.0K & VL-T5 & 98.5 & 98.5 & 100 & 100 & 68.5 & 99.5 & 100 & 21.5 & 100 & 99.0 & 54.5 & 99.5 & 86.6\\
            \midrule
            31K & T5 + RC (Ours) & 97.0 & 98.0 & 99.0 & 99.0 & 90.5 & 96.0 & 99.5 & 45.5 & 98.0 & 97.0 & 47.5 & 96.5 & 88.6\\
            31K & T5 & 96.0 & 99.0 & 99.5 & 100 & 98.0 & 97.0 & 96.0 & 21.5 & 100 & 95.5 & 42.0 & 99.5 & 87.0\\
            31K & VL-T5 & 96.5 & 97.0 & 99.5 & 99.5 & 69.0 & 94.5 & 95.0 & 21.0 & 99.5 & 96.5 & 42.0 & 100 & 84.2\\
            \midrule
            52K & T5 + RC (Ours) & 99.5 & 99.0 & 100 & 99.0 & 93.0 & 98.0 & 99.0 & 98.0 & 98.0 & 95.5 & 46.0 & 97.0 & 93.5\\
            52K & T5 & 96.5 & 96.0 & 99.5 & 100 & 97.5 & 98.5 & 97.0 & 17.5 & 100 & 97.0 & 38.5 & 100 & 86.5\\
            52K & VL-T5 & 96.0 & 98.5 & 99.5 & 100 & 96.5 & 95.5 & 95.5 & 21.0 & 100 & 98.0 & 41.5 & 100 & 86.8\\
            \midrule
            103K & T5 + RC (Ours) & 98.0 & 99.0 & 100 & 99.5 & 94.0 & 97.5 & 99.0 & 97.0 & 96.5 & 95.0 & 47.0 & 98.0 & 93.4\\
            103K & T5 & 99.0 & 97.5 & 100 & 99.5 & 96.5 & 96.0 & 95.0 & 15.5 & 100 & 95.5 & 43.5 & 99.5 & 86.5\\
            103K & VL-T5 & 98.0 & 97.0 & 100 & 99.5 & 96.0 & 97.0 & 96.5 & 84.5 & 100 & 99.5 & 41.0 & 99.5 & 92.4\\
            \bottomrule
        \end{tabular}
    }
\end{table}

\begin{table}[ht]
    \caption {Comparison of the performance of our method with different multimodal prompt encoder on L4 level generalization. All methods share the same amount of parameters 92M. Integers in the first row refer to indices of tasks defined in the VIMA paper~\citep{jiang2023vima}}\label{tab:prompt-encoder-result-L4}
    \centering
    {
        \begin{tabular}{l|l|cccc|c}
            \toprule
            Pretrain iter. & Method &  T8 & T10 & T13 & T14 & Overall\\
            \midrule
            0.0K & T5 + RC (Ours) & 97.0 & 0.0 & 0.0 & 99.5 & 49.1\\
            0.0K & T5 & 95.0 & 0.0 & 0.0 & 98.5 & 48.4\\
            0.0K & VL-T5 & 99.0 & 0.0 & 0.0 & 97.5 & 49.1\\
            \midrule
            31K & T5 + RC (Ours) & 97.5 & 12.5 & 0.0 & 98.5 & 52.1\\
            31K & T5 & 98.0 & 45.0 & 0.0 & 95.5 & 59.6\\
            31K & VL-T5 & 98.5 & 64.0 & 0.0 & 96.0 & 64.6\\
            \midrule
            52K & T5 + RC (Ours) & 98.0 & 40.5 & 0.0 & 96.5 & 58.8\\
            52K & T5 & 98.5 & 55.5 & 0.0 & 96.0 & 62.5\\
            52K & VL-T5 & 98.0 & 37.5 & 0.0 & 96.5 & 58.0\\
            \midrule
            103K & T5 + RC (Ours) & 97.5 & 41.0 & 1.0 & 97.0 & 59.1\\
            103K & T5 & 98.5 & 39.5 & 0.0 & 98.5 & 59.1\\
            103K & VL-T5 & 97.5 & 53.5 & 0.0 & 98.0 & 62.3\\
            \bottomrule
        \end{tabular}
    }
\end{table}

\begin{table}[ht]
    \caption {Comparison of the performance of our method with different model sizes ranging from 2M to 92M on L1 level generalization results. Integers in the first row refer to indices of tasks defined in the VIMA paper~\citep{jiang2023vima}}\label{tab:model-size-result-L1}
    \centering
    \resizebox{\linewidth}{!}{
        \begin{tabular}{l|l|ccccccccccccc|c}
            \toprule
            Model size. & Method  & T1 & T2 & T3 & T4 & T5 & T6 & T7 & T9 & T11 & T12 & T15 & T16 & T17 & Overall\\
            \midrule
            2M & Ours w/o Pretrain & 100 & 98.5 & 99.0 & 89.5 & 48.5 & 100 & 100 & 19.5 & 97.0 & 91.0 & 98.0 & 36.0 & 24.0 & 77.0\\
            2M & Ours & 99.5 & 99.0 & 97.5 & 99.0 & 67.5 & 100 & 99.5 & 18.5 & 91.5 & 93.0 & 99.0 & 38.0 & 64.5 & 82.0\\
            \midrule
            4M & Ours w/o Pretrain & 100 & 100 & 99.5 & 97.0 & 55.0 & 100 & 100 & 18.0 & 96.0 & 95.0 & 99.5 & 44.0 & 40.0 & 80.3\\
            4M & Ours & 100 & 100 & 86.5 & 99.0 & 63.5 & 99.5 & 100 & 20.5 & 92.0 & 95.5 & 98.0 & 83.5 & 57.0 & 84.2\\
            \midrule
            9M & Ours w/o Pretrain & 100 & 100 & 96.0 & 99.0 & 57.0 & 100 & 100 & 23.0 & 98.0 & 94.0 & 98.5 & 47.0 & 94.0 & 85.1\\
            9M & Ours & 100 & 100 & 99.0 & 99.0 & 87.0 & 100 & 100 & 19.0 & 100 & 95.5 & 98.5 & 92.5 & 97.0 & 91.3\\
            \midrule
            20M & Ours w/o Pretrain & 100 & 100 & 100 & 98.5 & 67.5 & 100 & 100 & 30.5 & 98.5 & 95.0 & 99.0 & 49.5 & 85.0 & 86.4\\
            20M & Ours & 100 & 100 & 100 & 97.0 & 90.0 & 100 & 99.5 & 19.0 & 100 & 94.0 & 99.5 & 93.5 & 97.5 & 91.5\\
            \midrule
            43M & Ours w/o Pretrain & 100 & 100 & 100 & 98.5 & 67.0 & 100 & 100 & 17.0 & 100 & 94.0 & 99.0 & 92.5 & 96.5 & 89.6\\
            43M & Ours & 99.5 & 100 & 99.5 & 95.5 & 89.0 & 97.5 & 100 & 100 & 100 & 94.5 & 96.0 & 94.5 & 96.5 & 97.1\\
            \midrule
            92M & Ours w/o Pretrain & 100 & 100 & 100 & 98.5 & 88.0 & 100 & 100 & 20.5 & 100 & 94.0 & 99.0 & 93.0 & 98.0 & 91.6\\
            92M & Ours & 98.5 & 100 & 100 & 99.5 & 94.0 & 100 & 100 & 100 & 100 & 94.0 & 95.5 & 94.0 & 96.5 & 97.8\\
             \bottomrule
        \end{tabular}
    }
\end{table}

\begin{table}[ht]
    \caption {Comparison of the performance of our method with different model sizes ranging from 2M to 92M on L2 level generalization results. Integers in the first row refer to indices of tasks defined in the VIMA paper~\citep{jiang2023vima}}\label{tab:model-size-result-L2}
    \centering
    \resizebox{\linewidth}{!}{
        \begin{tabular}{l|l|ccccccccccccc|c}
            \toprule
            Model size. & Method  & T1 & T2 & T3 & T4 & T5 & T6 & T7 & T9 & T11 & T12 & T15 & T16 & T17 & Overall\\
            \midrule
            2M & Ours w/o Pretrain & 95.5 & 84.5 & 99.0 & 87.0 & 50.0 & 96.5 & 91.0 & 21.0 & 97.0 & 91.0 & 88.0 & 33.5 & 11.5 & 72.7\\
            2M & Ours & 99.5 & 98.5 & 98.5 & 98.5 & 59.0 & 100 & 98.5 & 20.5 & 92.0 & 92.5 & 99.0 & 39.5 & 61.5 & 81.3\\
            \midrule
            4M & Ours w/o Pretrain & 99.0 & 98.5 & 100 & 97.0 & 55.0 & 99.5 & 98.5 & 21.0 & 96.5 & 95.5 & 97.0 & 44.0 & 35.0 & 79.7\\
            4M & Ours & 100 & 100 & 87.0 & 99.0 & 67.5 & 99.5 & 99.5 & 19.0 & 92.5 & 95.5 & 97.0 & 84.0 & 60.0 & 84.7\\
            \midrule
            9M & Ours w/o Pretrain & 100 & 100 & 96.5 & 98.5 & 58.0 & 99.5 & 99.0 & 25.5 & 97.5 & 94.5 & 94.5 & 47.0 & 88.0 & 84.5\\
            9M & Ours & 100 & 100 & 99.5 & 98.5 & 86.5 & 99.5 & 100 & 19.0 & 100 & 94.5 & 97.0 & 91.5 & 95.0 & 90.8\\
            \midrule
            20M & Ours w/o Pretrain & 100 & 100 & 100 & 98.5 & 72.0 & 100 & 100 & 29.5 & 98.0 & 95.5 & 99.0 & 46.0 & 83.5 & 86.3\\
            20M & Ours & 100 & 100 & 100 & 97.0 & 86.5 & 99.0 & 99.0 & 19.5 & 100 & 95.0 & 97.0 & 91.5 & 96.5 & 90.8\\
            \midrule
            43M & Ours w/o Pretrain & 100 & 100 & 100 & 98.5 & 72.5 & 100 & 100 & 18.5 & 100 & 93.5 & 99.5 & 92.0 & 96.0 & 90.0\\
            43M & Ours & 99.0 & 100 & 100 & 97.0 & 90.5 & 98.0 & 100 & 99.5 & 100 & 95.5 & 94.0 & 93.0 & 96.5 & 97.2\\
            \midrule
            92M & Ours w/o Pretrain & 100 & 100 & 100 & 99.0 & 87.0 & 100 & 100 & 23.5 & 100 & 94.0 & 99.5 & 92.0 & 98.0 & 91.8\\
            92M & Ours & 98.5 & 100 & 100 & 99.0 & 96.5 & 99.5 & 100 & 100 & 100 & 95.5 & 95.0 & 93.0 & 96.0 & 97.9\\

             \bottomrule
        \end{tabular}
    }
\end{table}

\begin{table}[ht]
    \caption {Comparison of the performance of our method with different model sizes ranging from 2M to 92M on L3 level generalization results. Integers in the first row refer to indices of tasks defined in the VIMA paper~\citep{jiang2023vima}}\label{tab:model-size-result-L3}
    \centering
    \resizebox{\linewidth}{!}{
        \begin{tabular}{l|l|cccccccccccc|c}
            \toprule
            Model size. & Method  & T1 & T2 & T3 & T4 & T5 & T6 & T7 & T9 & T11 & T15 & T16 & T17 & Overall\\
            \midrule
            2M & Ours w/o Pretrain & 97.0 & 91.0 & 100 & 92.5 & 46.0 & 96.5 & 95.5 & 15.5 & 95.5 & 95.0 & 36.0 & 8.0 & 72.4\\
            2M & Ours & 99.0 & 98.5 & 99.5 & 99.5 & 67.0 & 99.5 & 99.5 & 16.5 & 83.5 & 98.5 & 33.5 & 54.0 & 79.0\\
            \midrule
            4M & Ours w/o Pretrain & 98.5 & 98.0 & 100 & 94.5 & 53.5 & 95.0 & 99.0 & 17.0 & 99.0 & 87.5 & 47.0 & 5.5 & 74.5\\
            4M & Ours & 95.5 & 99.0 & 75.5 & 95.5 & 61.0 & 95.5 & 96.5 & 19.0 & 92.5 & 79.5 & 42.0 & 32.0 & 73.6\\
            \midrule
            9M & Ours w/o Pretrain & 93.5 & 97.5 & 96.0 & 100 & 64.0 & 95.0 & 96.0 & 17.5 & 97.5 & 85.0 & 43.0 & 44.0 & 77.4\\
            9M & Ours & 97.0 & 97.0 & 100 & 96.5 & 86.0 & 99.0 & 99.0 & 23.5 & 98.5 & 96.0 & 52.5 & 100 & 87.1\\
            \midrule
            20M & Ours w/o Pretrain & 99.5 & 100 & 100 & 100 & 71.5 & 100 & 100 & 26.0 & 98.5 & 98.5 & 43.5 & 87.5 & 85.4\\
            20M & Ours & 97.0 & 97.0 & 99.5 & 99.5 & 89.0 & 98.0 & 98.0 & 24.0 & 100 & 99.0 & 53.5 & 98.0 & 87.7\\
            \midrule
            43M & Ours w/o Pretrain & 99.5 & 100 & 100 & 100 & 74.0 & 100 & 99.5 & 25.0 & 100 & 99.5 & 54.0 & 99.0 & 87.5\\
            43M & Ours & 95.0 & 98.0 & 99.5 & 96.5 & 86.0 & 95.5 & 96.5 & 97.0 & 99.5 & 96.0 & 40.0 & 99.0 & 91.5\\
            \midrule
            92M & Ours w/o Pretrain & 99.5 & 100 & 100 & 100 & 90.0 & 100 & 100 & 20.5 & 100 & 99.5 & 50.5 & 99.5 & 88.3\\
            92M & Ours & 98.0 & 99.0 & 100 & 99.5 & 94.0 & 97.5 & 99.0 & 97.0 & 96.5 & 95.0 & 47.0 & 98.0 & 93.4\\
            \bottomrule
        \end{tabular}
    }
\end{table}

\begin{table}[ht]
    \caption {Comparison of the performance of our method with different model sizes ranging from 2M to 92M on L4 level generalization results. Integers in the first row refer to indices of tasks defined in the VIMA paper~\citep{jiang2023vima}}\label{tab:model-size-result-L4}
    \centering
    {
        \begin{tabular}{l|l|cccc|c}
            \toprule
            Model size. & Method &  T8 & T10 & T13 & T14 & Overall\\
            \midrule
            2M & Ours w/o Pretrain & 78.5 & 0.0 & 0.0 & 95.5 & 43.5\\
            2M & Ours & 47.5 & 35.5 & 0.5 & 97.5 & 45.2\\
            \midrule
            4M & Ours w/o Pretrain & 99.5 & 0.0 & 0.0 & 95.5 & 48.8\\
            4M & Ours & 96.0 & 0.5 & 0.0 & 92.5 & 47.2\\
            \midrule
            9M & Ours w/o Pretrain & 96.5 & 1.0 & 0.0 & 95.0 & 48.1\\
            9M & Ours & 99.5 & 15.5 & 0.5 & 98.0 & 53.4\\
            \midrule
            20M & Ours w/o Pretrain & 99.0 & 0.0 & 0.0 & 99.0 & 49.5\\
            20M & Ours & 97.0 & 22.0 & 0.0 & 95.5 & 53.6\\
            \midrule
            43M & Ours w/o Pretrain & 99.0 & 0.0 & 0.0 & 98.5 & 49.4\\
            43M & Ours & 95.5 & 6.0 & 0.0 & 96.0 & 49.4\\
            \midrule
            92M & Ours w/o Pretrain & 97.0 & 0.0 & 0.0 & 99.5 & 49.1\\
            92M & Ours & 97.5 & 41.0 & 1.0 & 97.0 & 59.1\\
            \bottomrule
        \end{tabular}
    }
\end{table}

\begin{table}[ht]
    \caption {Comparison of the performance of our method with different scales of training data on L1 level generalization results. Integers in the first row refer to indices of tasks defined in the VIMA paper~\citep{jiang2023vima}}\label{tab:data-scale-result-L1}
    \centering
    \resizebox{\linewidth}{!}{
        \begin{tabular}{l|l|ccccccccccccc|c}
            \toprule
            Data Size & Method  & T1 & T2 & T3 & T4 & T5 & T6 & T7 & T9 & T11 & T12 & T15 & T16 & T17 & Overall\\
            \midrule
            10\% & Ours w/o Pretrain & 100 & 98.5 & 96.5 & 97.0 & 74.0 & 97.5 & 100 & 19.0 & 100 & 93.0 & 93.0 & 88.0 & 93.0 & 88.4\\
            10\% & Ours & 100 & 99.5 & 96.5 & 89.0 & 65.5 & 98.0 & 98.5 & 73.5 & 97.5 & 93.5 & 92.0 & 89.0 & 93.0 & 91.2\\
            \midrule
            50\% & Ours w/o Pretrain & 100 & 99.5 & 97.5 & 98.0 & 74.0 & 99.5 & 99.5 & 20.0 & 100 & 92.5 & 98.0 & 82.0 & 91.0 & 88.6\\
            50\% & Ours & 100 & 99.5 & 98.5 & 98.0 & 86.5 & 99.5 & 99.5 & 98.5 & 100 & 93.5 & 96.5 & 95.5 & 96.5 & 97.1\\
            \midrule
            100\% & Ours w/o Pretrain & 100 & 100 & 100 & 98.5 & 88.0 & 100 & 100 & 20.5 & 100 & 94.0 & 99.0 & 93.0 & 98.0 & 91.6\\
            100\% & Ours & 98.5 & 100 & 100 & 99.5 & 94.0 & 100 & 100 & 100 & 100 & 94.0 & 95.5 & 94.0 & 96.5 & 97.8\\
             \bottomrule
        \end{tabular}
    }
\end{table}

\begin{table}[ht]
    \caption {Comparison of the performance of our method with different scales of training data on L2 level generalization results. Integers in the first row refer to indices of tasks defined in the VIMA paper~\citep{jiang2023vima}}\label{tab:data-scale-result-L2}
    \centering
    \resizebox{\linewidth}{!}{
        \begin{tabular}{l|l|ccccccccccccc|c}
            \toprule
            Data Size & Method  & T1 & T2 & T3 & T4 & T5 & T6 & T7 & T9 & T11 & T12 & T15 & T16 & T17 & Overall\\
            \midrule
            10\% & Ours w/o Pretrain & 99.5 & 99.0 & 97.0 & 95.5 & 72.5 & 97.5 & 99.5 & 20.5 & 98.5 & 93.0 & 91.5 & 88.5 & 91.0 & 88.0\\
            10\% & Ours & 99.0 & 99.5 & 94.5 & 90.0 & 62.5 & 98.5 & 99.0 & 77.5 & 98.5 & 94.0 & 90.0 & 87.0 & 89.0 & 90.7\\
            \midrule
            50\% & Ours w/o Pretrain & 98.5 & 100 & 97.0 & 98.0 & 72.0 & 99.5 & 99.5 & 16.5 & 100 & 91.5 & 97.5 & 84.5 & 88.5 & 87.9\\
            50\% & Ours & 100 & 100 & 99.0 & 97.5 & 88.5 & 99.0 & 99.0 & 98.5 & 100 & 95.0 & 96.0 & 95.5 & 96.5 & 97.3\\
            \midrule
            100\% & Ours w/o Pretrain & 100 & 100 & 100 & 99.0 & 87.0 & 100 & 100 & 23.5 & 100 & 94.0 & 99.5 & 92.0 & 98.0 & 91.8\\
            100\% & Ours & 98.5 & 100 & 100 & 99.0 & 96.5 & 99.5 & 100 & 100 & 100 & 95.5 & 95.0 & 93.0 & 96.0 & 97.9\\
             \bottomrule
        \end{tabular}
    }
\end{table}

\begin{table}[ht]
    \caption {Comparison of the performance of our method with different scales of training data on L3 level generalization results. Integers in the first row refer to indices of tasks defined in the VIMA paper~\citep{jiang2023vima}}\label{tab:data-scale-result-L3}
    \centering
    \resizebox{\linewidth}{!}{
        \begin{tabular}{l|l|cccccccccccc|c}
            \toprule
            Data Size & Method  & T1 & T2 & T3 & T4 & T5 & T6 & T7 & T9 & T11 & T15 & T16 & T17 & Overall\\
            \midrule
            10\% & Ours w/o Pretrain & 98.0 & 97.0 & 97.5 & 98.5 & 74.5 & 97.5 & 99.0 & 18.0 & 100 & 96.5 & 53.5 & 99.0 & 85.8\\
            10\% & Ours & 90.0 & 93.5 & 98.5 & 93.0 & 71.0 & 90.0 & 90.5 & 56.0 & 90.0 & 83.0 & 52.0 & 20.5 & 77.3\\
            \midrule
            50\% & Ours w/o Pretrain & 97.5 & 99.5 & 99.0 & 99.5 & 70.5 & 98.5 & 99.0 & 19.0 & 100 & 97.0 & 57.5 & 93.5 & 85.9\\
            50\% & Ours & 97.0 & 97.5 & 99.0 & 99.5 & 86.0 & 97.5 & 96.5 & 95.5 & 98.0 & 97.0 & 47.5 & 100 & 92.6\\
            \midrule
            100\% & Ours w/o Pretrain & 99.5 & 100 & 100 & 100 & 90.0 & 100 & 100 & 20.5 & 100 & 99.5 & 50.5 & 99.5 & 88.3\\
            100\% & Ours & 98.0 & 99.0 & 100 & 99.5 & 94.0 & 97.5 & 99.0 & 97.0 & 96.5 & 95.0 & 47.0 & 98.0 & 93.4\\
            \bottomrule
        \end{tabular}
    }
\end{table}

\begin{table}[ht]
    \caption {Comparison of the performance of our method with different scales of training data on L4 level generalization results. Integers in the first row refer to indices of tasks defined in the VIMA paper~\citep{jiang2023vima}}\label{tab:data-scale-result-L4}
    \centering
    {
        \begin{tabular}{l|l|cccc|c}
            \toprule
            Data Size & Method &  T8 & T10 & T13 & T14 & Overall\\
            \midrule
            10\% & Ours w/o Pretrain & 92.0 & 0.0 & 0.0 & 94.5 & 46.6\\
            10\% & Ours & 91.0 & 39.0 & 0.0 & 88.0 & 54.5\\
            \midrule
            50\% & Ours w/o Pretrain & 91.5 & 0.0 & 0.0 & 97.0 & 47.1\\
            50\% & Ours & 95.0 & 12.5 & 0.0 & 96.0 & 50.9\\
            \midrule
            100\% & Ours w/o Pretrain & 97.0 & 0.0 & 0.0 & 99.5 & 49.1\\
            100\% & Ours & 97.5 & 41.0 & 1.0 & 97.0 & 59.1\\
            \bottomrule
        \end{tabular}
    }
\end{table}

\clearpage
\section{Pseudo-codes \& Training Details}\label{sec:pseudo-codes}
\begin{algorithm}[th]
    \caption{Robot Control with multimodal prompts through pretraining and multitask FT}\label{alg}
    {\bf Input}: Dataset $\mathcal{D} = \{\zeta_1, \zeta_2, \ldots\}$, policy parameter $\theta$, number of pretraining iterations $N_{\text{pretrain}}$, number of multi-task imitation finetuning iterations $N_{\text{FT}}$
    \begin{algorithmic}[1]
    \FOR{$i= 1, \ldots, N_{\text{pretrain}}$}
        \STATE Sample a mini-batch $\mathcal{B}$ from $\mathcal{D}$
        \STATE Minimize $L_{\text{pretrain}}(\theta)$ defined in Eq. \ref{eq:pretrain_loss} on $\mathcal{B}$
    \ENDFOR
    \FOR{$i= 1, \ldots, N_{\text{FT}}$}
        \STATE Sample a mini-batch $\mathcal{B}$ from $\mathcal{D}$
        \STATE Minimize $L_{\text{Imitatation}}(\theta)$ defined in Eq. \ref{eq:imitate_loss} on $\mathcal{B}$
    \ENDFOR
    \end{algorithmic}
\end{algorithm}

\begin{figure}[ht]
     \centering
     \includegraphics[width=\textwidth]{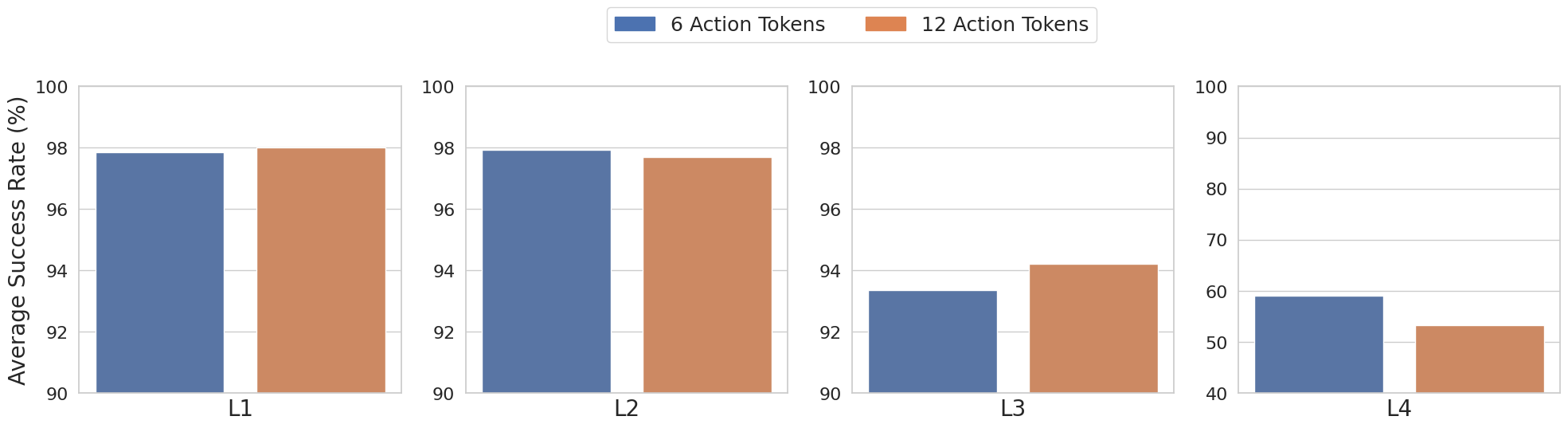}
    \caption{Ablation on the number of action tokens.}
    \label{fig:ablate_act_token_num}
\end{figure}

Algorithm \ref{alg} presents the pseudo-codes for the training pipeline, which includes a pretraining phase and a multi-task FT phase. We set our training HP following the recipe provided by VIMA, which open-sourced its policy architectures without providing the training codes.
We conduct our experiments on cluster nodes, each with 8 NVIDIA-A10G. \autoref{tab:main-hp} presents the HP for our training pipeline. As we build our policy based on the VIMA Policy, we refer interested readers to Tables 2 and 3 in Appendix C of VIMA paper~\citep{jiang2023vima} for all model parameters.

Additionally, the action space $\mathcal{A}$ includes initial pose $\mathcal{T}_{\text{initial}}\in\mathcal{R}^6$ and $\mathcal{T}_{\text{target}}\in\mathcal{R}^6$. Each pose is a 6-dimension vector with 2 for xy position and 4 for rotation represented in quaternion. Since the VIMA-BENCH focuses on tabletop manipulation, the rotation quaternion of $\mathcal{T}_{\text{initial}}$ is always a constant vector. So is the first two dimensions of the rotation quaternion of $\mathcal{T}_{\text{initial}}$. Therefore, we only tokenize the other 6 action dimensions to improve computational efficiency. Thus, each action worth 6 tokens. Moreover, we conduct an ablation study to show that this choice will not affect the task success rate. As shown in \autoref{fig:ablate_act_token_num}, modeling each of the 12 action dimensions as a single token achieves almost the same performance as modeling the 6 active action dimensions.
\begin{table}[ht]
    \caption {Hyper-parameters for our training pipeline}\label{tab:main-hp}
    \centering
        \begin{tabular}{lll}
            \toprule
            Phase & Hyperparameter &  Value\\
            \midrule
                   & Learning Rate (LR) & 1e-4\\
                   & Minimum LR & 1e-7\\
                   & Warmup Steps & 7K \\
            Shared & Weight Decay & 0 \\
                   & Dropout & 0.1 \\
                   & Gradient Clip Threshold & 1.0 \\
                   & Optimizer & AdamW~\citep{loshchilov2017decoupled} \\
                   & Batch Size & 128 \\
                   & Iterations per epochs & 5158 \\
            \midrule
            Pretrain & Training epochs & 20\\
                     & Training iterations $N_{\text{pretrain}}$ & 20 $\times$ Iterations per epochs = 103160 \\
                     & LR Cosine Annealing Steps & $N_{\text{pretrain}}$ - Warmup Steps = 96160\\

            \midrule
            Finetune & LR Cosine Annealing Steps & 17K\\
                     & Training epochs & 10\\
                     & Training iterations $N_{\text{FT}}$  & 10 $\times$ Iterations per epochs = 51580 \\
            \bottomrule
        \end{tabular}
\end{table}
\clearpage

\section{Details of Evaluating the In-context Learning Ability}\label{sec:appen-in-context-exp}
\begin{figure*}[ht]
  \makebox[\textwidth]
    {
        \begin{subfigure}{0.26\paperwidth}
            \center
            \includegraphics[width=\linewidth]{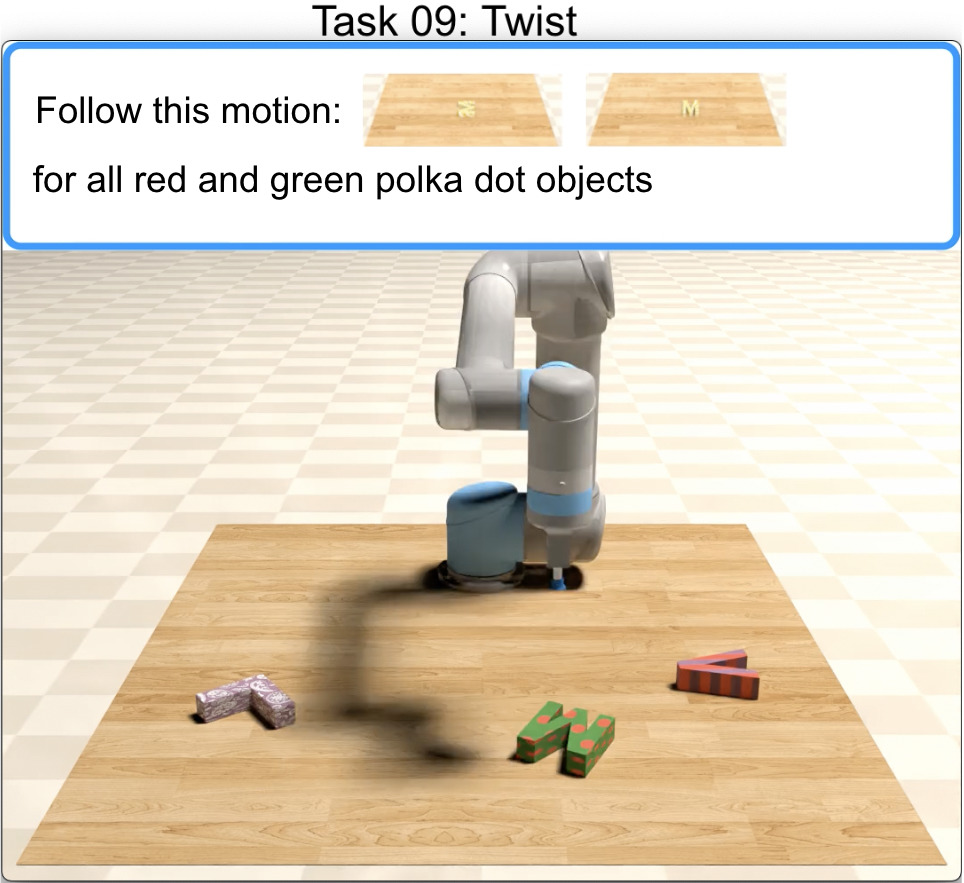}
        \end{subfigure}
        \begin{subfigure}{0.26\paperwidth}
            \center
            \includegraphics[width=\linewidth]{figures/follow_motion.jpg}
        \end{subfigure}

        \begin{subfigure}{0.26\paperwidth}
            \center
            \includegraphics[width=\linewidth]{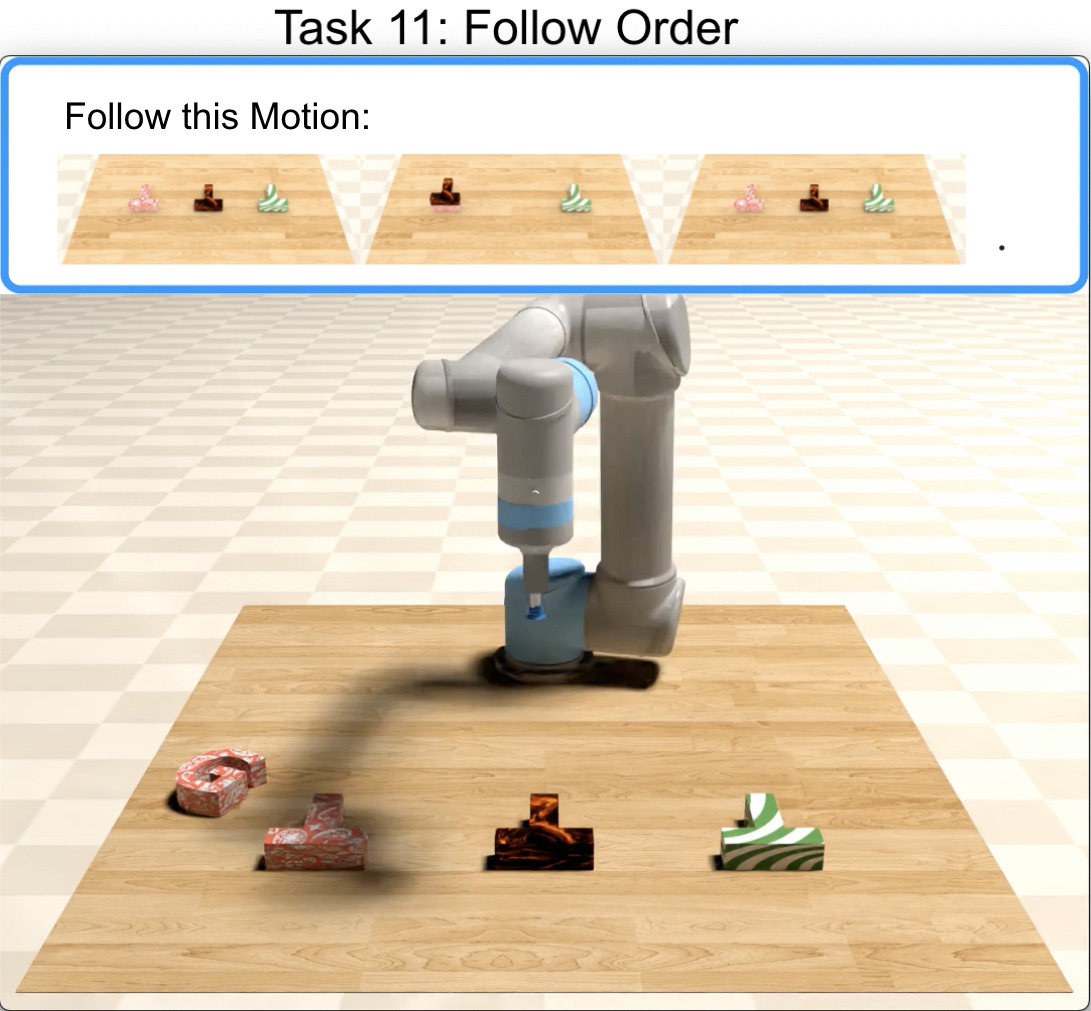}
        \end{subfigure}
    }
  \caption{The new set of L4 tasks with in-context examples and modified prompts.}
\label{fig:modified-icl-tasks}
\end{figure*}

We provide training details for the experiments conducted in Sec. \ref{sec:in-context-exp} by introducing the data augmentation strategies, pretraining, our \emph{Modified FT}, and how we edit the task prompt for Task 09 (\emph{Twist}) and Task 10 (\emph{Follow Order}). Moreover, the L1, L2 and L3 success rate in this settings are given by $97.6\%$, $97.7\%$, and $93.0\%$, respectively.

\textbf{Data Augmentation} To improve the generalizability of the pretrained model, we randomly apply the standard random data augmentation techniques, including Color Jitter and Gray Scale~\citep{he2020momentum} to the prompt images. Since we adopt an object-centric representation, we randomly shift the bounding box location for all objects in the whole trajectory with the same constant value. Note that we only augment the prompt images without modifying the observation images.

\textbf{Pretraining} We empirically find that further dividing the pretraining phase into two steps can improve the performance. We first pretrain a policy for 20 epochs and only extract the object encoder from it. Next, we use the pretrained object encoder to initialize another policy and pretrain it for 5 epochs. And the FT phase remains unchanged.

\textbf{Modified FT} To improve the model's ability to understand both visual and textual object descriptions, we randomly replace the object images in the multimodal prompts with text descriptions during multi-task FT. For example, the task prompt for \emph{Follow Motion} in \autoref{fig:modified-icl-tasks} can be rephrased as
\begin{mdframed}
    Follow this motion for the white and purple striped V: $\{\text{frame}_1\}$, $\{\text{frame}_2\}$, $\{\text{frame}_3\}$.
\end{mdframed}
Note that only object images will be converted into text descriptions. Images depicted the scene, e.g., $\text{frame}_1$, $\text{frame}_2$, $\text{frame}_3$, will never be converted to text. We randomly apply this operation to the task prompt of the pretraining tasks during the FT phase.

\textbf{Edit Prompts} As shown in \autoref{fig:modified-icl-tasks}, we modify the task prompt for both \emph{Twist} and \emph{Follow order} to make them similar to the pretraining prompts. Specifically, the task prompt for \emph{Twist} is modified as below
\begin{mdframed}
\textbf{Original}: ``Twist" is defined as rotating object a specific angle. For examples: From $\{\text{before\_twist}_1\}$ to $\{\text{after\_twist}_1\}$. From $\{\text{before\_twist}_2\}$ to $\{\text{after\_twist}_2\}$. From $\{\text{before\_twist}_3\}$ to $\{\text{after\_twist}_3\}$. Now twist all [TEXT OBJ DESCRIPTION] objects.

\textbf{Modified}: Follow this motion: $\{\text{before\_twist}_1\}$ to $\{\text{after\_twist}_1\}$ for all [TEXT OBJ DESCRIPTION] objects.
\end{mdframed}
Similarly, the task prompt for \emph{Follow Order} is modified as below:
\begin{mdframed}
\textbf{Original}: Stack objects in this order $\{\text{frame}_1\}$, $\{\text{frame}_2\}$, $\{\text{frame}_3\}$.

\textbf{Modified}: Follow this motion: $\{\text{frame}_1\}$, $\{\text{frame}_2\}$, $\{\text{frame}_3\}$.

\end{mdframed}

\blue{Although the modified prompt seems to be controlled at the first glance, we emphasize that any multimodal prompt with in-context examples can be paraphrased to fit into the templates above. And thus, we did not consider diversifying the prompt language in our experiments.}

\clearpage
\section{Additional L4 Unseen Tasks with In-context Examples}\label{sec:add_icl_exp}
We augment the L4 task suite of VIMA-BENCH by designing 4 new tasks with in-context examples provided in the prompt. These tasks are within the \emph{One-shot Video Imitation} category of VIMA-BENCH (Appendix B.4, ~\cite{jiang2023vima}). Next, we will first provide the task definitions. Then, we take our policy that is trained on the full data of VIMA-BENCH and evaluate it on these tasks. Notably, we never use trajectories collected from these tasks to train our policy.

\begin{figure*}[h]
  \makebox[\textwidth]
    {
        \begin{subfigure}{0.32\paperwidth}
            \center
            \includegraphics[width=\linewidth]{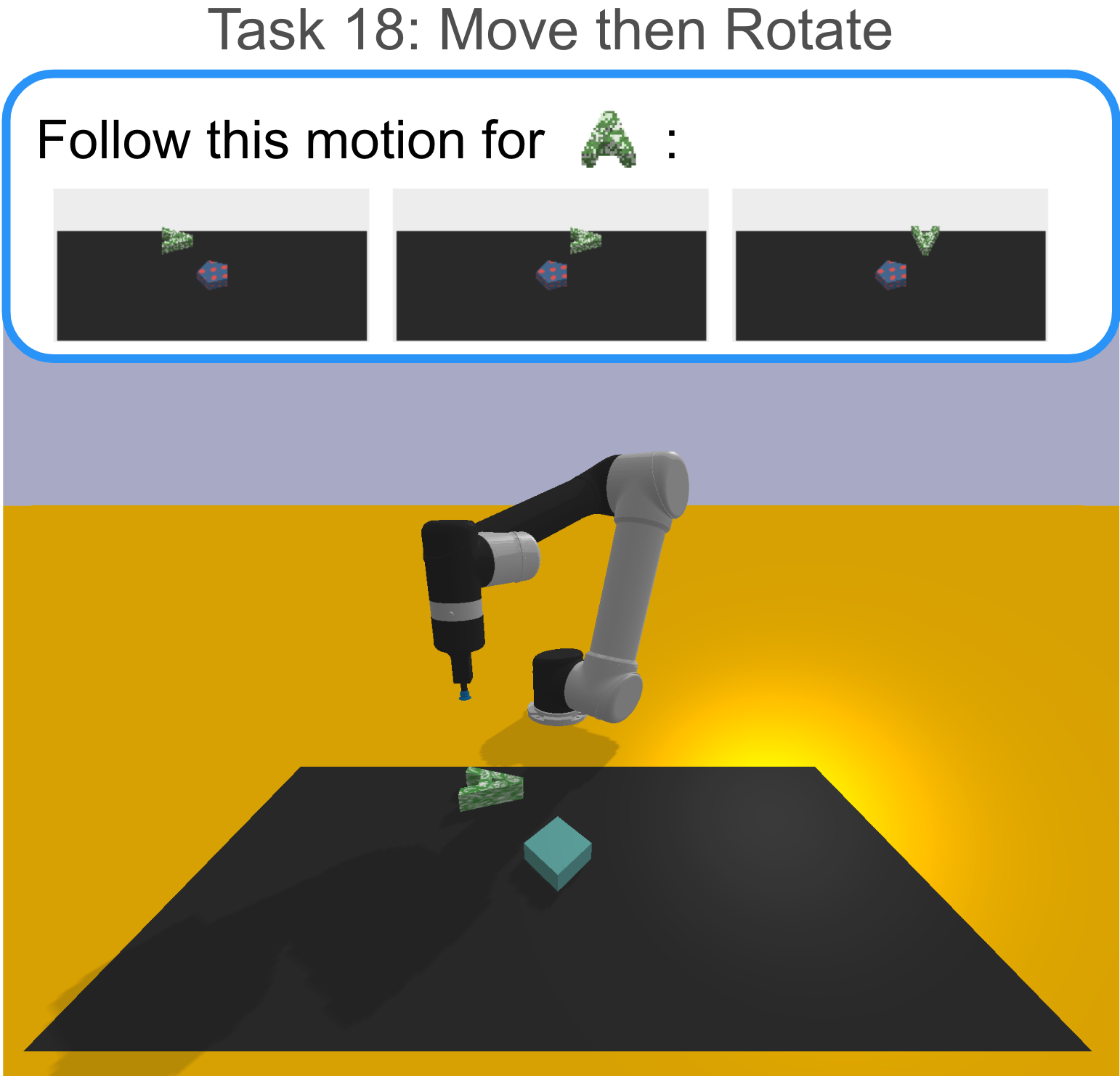}
        \end{subfigure}
        \begin{subfigure}{0.32\paperwidth}
            \center
            \includegraphics[width=\linewidth]{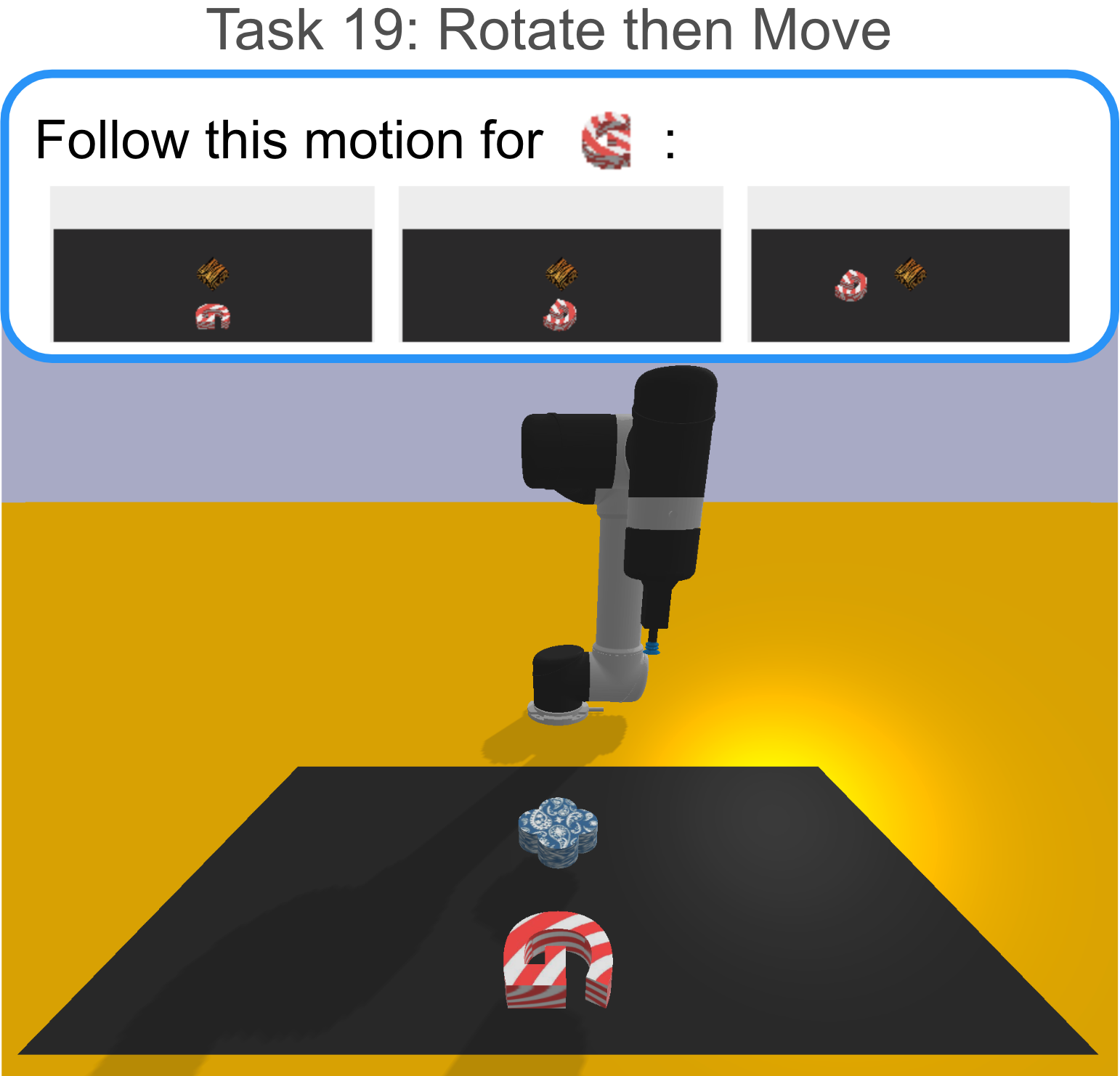}
        \end{subfigure}
    }
    \makebox[\textwidth]
    {
        \begin{subfigure}{0.32\paperwidth}
            \center
            \includegraphics[width=\linewidth]{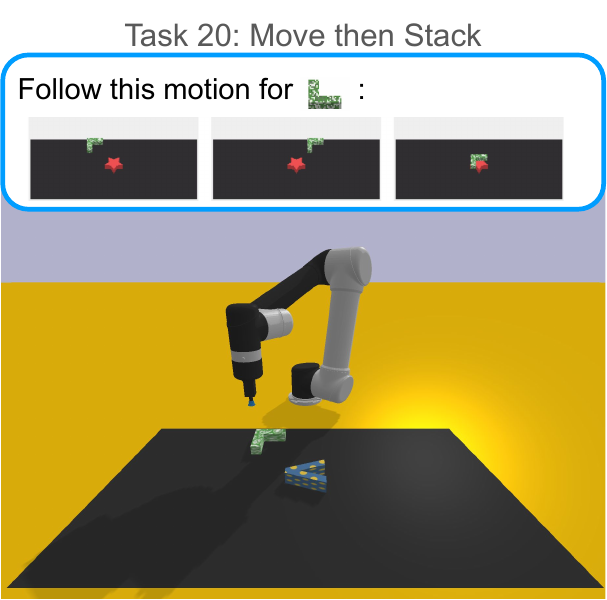}
        \end{subfigure}
        \begin{subfigure}{0.32\paperwidth}
            \center
            \includegraphics[width=\linewidth]{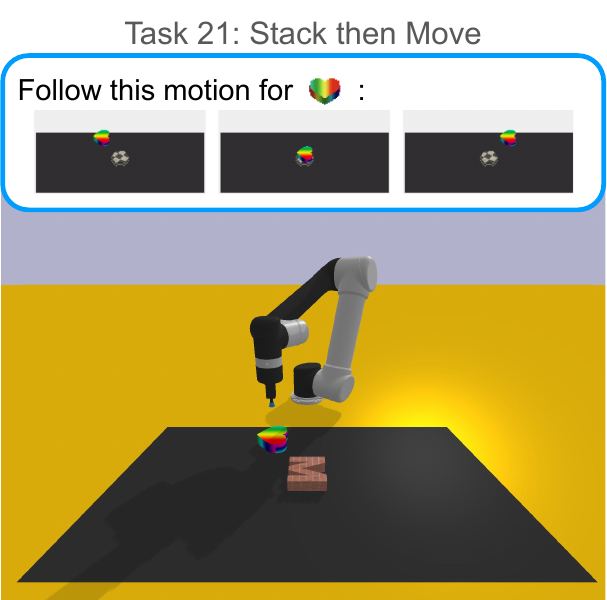}
        \end{subfigure}
    }
  \caption{Task samples from our designed tasks. Each task is paired with in-context demonstration in the prompt.
  }
\label{fig:add-icl-tasks}
\end{figure*}

\subsection{Task Definition}\label{sec:task_def}

To evaluate the in-context learning ability of a policy, we design four tasks to incorporate a demonstration trajectory in the task prompt. Specifically, these tasks share the same prompt template
\begin{mdframed}
    Follow this motion for  $\{\text{target object}\}$: $\{\text{frame}_1\}$, $\{\text{frame}_2\}$, $\{\text{frame}_3\}$.
\end{mdframed}
Note that we did not inject language variety to the task prompt, \textbf{as we can always paraphrase the task prompt to the unified prompt defined above given demonstration trajectory}.

Image placeholder \{target object\} is the target object to be manipulated and $\{\{\text{frame}_i\}\}$ a set of workspace-like scene placeholders to represent a video trajectory. Distractor objects are spawned at the center of the workspace and the prompt video. However, the distractor in the workspace is different from the distractor in the prompt video. The initial position of the target object matches that in $\{\text{frame}_1\}$. 

\textbf{Task 18: Move then Rotate}. The robot should first move the target object to a specific location and then rotate the target object by a certain degree, according to the demonstration trajectory.

\textbf{Task 19: Rotate then Move}. The robot should first rotate the target object by a certain degree and then move the target object to a specific location according to the demonstration trajectory.

\textbf{Task 20: Move then Stack}. The robot should first move the target object to a specific location and then stack the target object on the distractor according to the demonstration trajectory.

\textbf{Task 21: Stack then Move}. The robot should first stack the target object on the distractor and then move the target object to a specific location according to the demonstration trajectory.

\subsection{Experimental Results on the Unseen Tasks}
\begin{table}[h]
\centering
\caption{Evaluating the in-context learning capability of the learned model on the four unseen tasks proposed in Appendix \ref{sec:task_def}. All policies are trained on the full data of VIMA-BENCH.}
\begin{tabular}{l|ccccc}
\toprule
\textbf{Method}                  & \textbf{Task 18} & \textbf{Task 19} & \textbf{Task 20} & \textbf{Task 21} & \textbf{Overall} \\
\midrule
Our Method                       & $\mathbf{13.5\%}$                  & $\mathbf{14.5\%}$                    & $\mathbf{22.0\%}$                   & $\mathbf{10.0\%}$                   & $\mathbf{15.0\%}$                        \\
\midrule
Our Method w/  Pretrain Only      & 4.5\%                     & 0.5\%                     & 20.5\%                   & 1.5\%                    & 5.3\%                         \\
\midrule
Our Method w/o  Pretrain          & 0.0\%                     & 0.0\%                     & 0.0\%                    & 0.5\%                    & 0.1\%                         \\
\midrule
VIMA                             & 0.0\%                     & 0.0\%                     & 0.0\%                    & 0.5\%                    & 0.1\%                         \\
\bottomrule
\end{tabular}
\label{table:add_icl}
\end{table}
We take policies trained on the full VIMA-BENCH data and directly compare their performance on these four new tasks. As shown in \autoref{table:add_icl}, Our Method significantly outperforms the baseline methods. On the other hand, the VIMA policy struggles to perform well on these tasks, showing its inability to learn from the in-context demonstration. Moreover, comparing the performance of Our Method with Our Method w/ Pretrain Only, we can conclude that our two-stage training pipeline produces a better in-context learner. 

\clearpage
\section{Additional Experimental Results}\label{sec:add-exp}

\subsection{Our Policy Can Tackle Pure Language Task Prompts}
In this section, we show that our policy trained with task prompts interleaved image and text can also tackle pure language task prompts. We evaluate our policy derived with \emph{Modified FT} (Sec. \ref{sec:in-context-exp}) and select Task 1, 3, 16, and 17 for evaluation. These four tasks only contain object images in their task prompts, and thus, we can easily replace the object images with text descriptions ("the {obj\_colr} {obj\_name}", e.g., the {white and purple striped} {V}). The results are given in \autoref{tab:pure-text}:

\begin{table}[h]
    \centering
    \begin{tabular}{c|c|ccccc}
        \toprule
         Level&  Prompt type&  T1&  T3&  T16&  T17& \\
        \midrule
         L1&  Multi-modal&  100\%&  100\%&  92\%&  97.5\%& \\
         L1&  Pure text&  100\%&  100\%&  92\%&  97\%& \\
         \midrule
         L2&  Multi-modal&  100\%&  99.5\%&  92\%&  96.5\%& \\
         L2&  Pure text&  100\%&  99.5\%&  91.5\%&  97\%& \\
         \midrule
         L3&  Multi-modal&  100\%&  99.5\%&  92\%&  96.5\%& \\
         L3&  Pure text&  100\%&  99.5\%&  91.5\%&  97\%& \\
        \bottomrule
    \end{tabular}
    \caption{Our policy trained with vision-language task prompts can also tackle pure language task prompts.}
    \label{tab:pure-text}
\end{table}

\subsection{Training Baseline Methods for Extra Gradient Steps}\label{sec:more-iter-baseline}
In Sec.~\ref{sec:exp-vima}, we compare Our Method w/ Pretrain with the baseline methods trained with pure imitation learning loss, including Our Method w/o Pretrain and the VIMA policy. Due to the pretraining phase, Our Method w/ Pretrain is trained with more gradient steps than these baseline methods. In this section, we allow baseline methods to be trained for longer on the imitation learning loss. Specifically, We continue training the policy derived from Our Method w/o Pretrain and the VIMA policy with multi-task imitation learning loss for another 103K gradient steps and compare their performance with Our Method w/ Pretrain again. Now, all three methods train for the same 155K gradient steps. As shown in the \autoref{tab:more-iter-baseline}, Our Method w/o Pretrain still significantly outperforms the two baseline methods.

\begin{table}[h]
    \centering
    \begin{tabular}{l|cccc}
        \toprule
            155K gradient steps&  L1&  L2&  L3& L4\\
        \midrule
            VIMA&  84.3\%&  84.1\%&  80.2\%& 48.4\%\\
            Our Method w/o Pretrain&  92.2\%&  91.8\%&  87.2\%& 48.6\%\\
            Our Method w/ Pretrain &  \textbf{97.8\%} &  \textbf{97.9\%} &  \textbf{93.4\%} & \textbf{59.1\%} \\
        \bottomrule
    \end{tabular}
    \caption{Comparison between Our Method w/ Pretrain and the baseline methods trained with pure imitation learning loss. All methods train for the same 155K iterations. Our Method w/ Pretrain significantly outperforms the two baseline methods.}
    \label{tab:more-iter-baseline}
\end{table}

\subsection{Can VL-T5 Close the Performance Gap with Even More Gradient Steps?}\label{sec:more-iter-vl-t5}

In Sec.~\ref{sec:ablation}, we compare the performance between our multimodal prompt encoder and VL-T5 by pregraining for 103K gradient steps. In this section, we further pretrain with the VL-T5 for another 51.6K gradient steps, resulting in 155K pretraining steps in total. As shown in \autoref{tab:more-iter-vl-t5}, these extra pretraining steps lead to performance degradation with VL-T5. Conversely, this performance degradation does not happen with our multimodal prompt encoder (T5 + RC).
\begin{table}[h]
    \centering
    \begin{tabular}{l|c|cccc}
    \toprule
    & Pretraining steps & L1 & L2 & L3 & L4 \\
    \midrule
    VL-T5 & 103K & 94.3 \% & 94.1 \% & 92.4 \% & 62.3 \% \\
    T5 + RC (Ours) & 103K & 97.8 \% & 97.9 \% & 93.4 \% & 59.1 \% \\
    \midrule
    VL-T5 & 155K & 91.5 \% & 91.4 \% & 88.0 \% & 56.8 \% \\
    T5 + RC (Ours) & 155K & 97.7 \% & 97.5 \% & 94.3 \% & 56.5 \% \\
    \bottomrule
    \end{tabular}
    \caption{Pretraining with the VL-T5 for 155K gradient steps degrades performance compared to 103K steps.}
    \label{tab:more-iter-vl-t5}
\end{table}

\end{document}